\documentclass[11pt]{article}

\usepackage[final]{acl}

\usepackage{placeins}
\usepackage{times}
\usepackage{latexsym}
\usepackage{amssymb,mathtools}
\usepackage{booktabs,multirow,array}
\usepackage{graphicx}
\usepackage{subcaption}
\usepackage{xcolor}
\usepackage{microtype}
\usepackage{xspace}
\usepackage{enumitem}
\usepackage{algorithm}
\usepackage{algpseudocode}
\usepackage{url}
\usepackage{hyperref}
\usepackage[nameinlink,capitalise]{cleveref}
\usepackage{graphicx}  
\usepackage{enumitem}  
\usepackage{wrapfig}
\usepackage{ulem}
\usepackage{stfloats}   
\usepackage{capt-of}   

\usepackage{colortbl}

\providecommand{\tb}{\textsc{TB-2}\xspace}
\providecommand{\terminalbench}{\textsc{TerminalBench-2}\xspace}
\providecommand{\irs}{\textsc{IRS}\xspace}
\providecommand{\appworld}{\textsc{AppWorld}\xspace}
\providecommand{\scienceworld}{\textsc{ScienceWorld}\xspace}

\newcommand{\sj}[1]{\textcolor{orange}{SJ: #1}}

\definecolor{SkillLoss}{RGB}{250,232,232}
\definecolor{SkillImprove}{RGB}{235,246,238}
\definecolor{SkillNegText}{RGB}{160,52,52}

\newcommand{\skillpair}[3]{%
  #1$\!\rightarrow\!$#2\,%
  \textbf{\ensuremath{(#3)}}%
}
\usepackage{xcolor}
\usepackage[most]{tcolorbox}
\definecolor{procframe}{HTML}{3E6FA8}
\definecolor{cacheframe}{HTML}{C9822B}
\newtcolorbox{promptbox}[2]{%
  enhanced, colback=#2!4, colframe=#2, boxrule=0.6pt, arc=2pt,
  left=5pt, right=5pt, top=3pt, bottom=3pt,
  fonttitle=\bfseries\footnotesize, coltitle=white,
  title={#1}, toptitle=1pt, bottomtitle=1pt,
  fontupper=\footnotesize, before skip=0pt, after skip=0pt}

\usepackage{array}
\newtcolorbox{examplebox}[1]{%
  enhanced, colback=gray!3, colframe=gray!55, boxrule=0.5pt, arc=2pt,
  left=4pt, right=4pt, top=2pt, bottom=2pt,
  fonttitle=\bfseries\footnotesize, coltitle=black, colbacktitle=gray!15,
  title={#1}, fontupper=\scriptsize\raggedright,
  before skip=0pt, after skip=0pt, equal height group=ex}

\usepackage{pifont}
\tcbuselibrary{raster}
\definecolor{wzero}{HTML}{68A8D8}      
\definecolor{wzeroDark}{HTML}{3880C0}
\definecolor{wzeroLight}{HTML}{E8F0F8}
\definecolor{wone}{HTML}{F0C030}       
\definecolor{woneDark}{HTML}{A07E00}
\definecolor{woneLight}{HTML}{FBF3D3}
\definecolor{wmaint}{HTML}{68A848}     
\definecolor{wmaintDark}{HTML}{387018}
\definecolor{inkgray}{HTML}{405060}
\newtcolorbox{driftcard}[2]{%
  enhanced, colback=white, colframe=inkgray!30, boxrule=0.5pt, arc=3pt,
  left=5pt, right=5pt, top=4pt, bottom=4pt,
  title={\textbf{#1}}, fonttitle=\footnotesize, coltitle=inkgray,
  colbacktitle=inkgray!5, titlerule=0.3pt, toptitle=2pt, bottomtitle=2pt,
  after title={\hfill{\scriptsize\itshape\color{inkgray!80}#2}},
  fontupper=\scriptsize\raggedright, before skip=0pt, after skip=0pt,
  equal height group=driftcards}
\newcommand{\envpair}[2]{%
  \begin{tcbraster}[raster columns=2, raster column skip=6pt, raster equal height,
    raster before skip=1pt, raster after skip=4pt]
  \begin{tcolorbox}[enhanced, colback=wzeroLight, colframe=wzero, boxrule=0.5pt,
    arc=2pt, left=3pt, right=3pt, top=1.5pt, bottom=1.5pt,
    fontupper=\scriptsize\raggedright]
    {\color{wzeroDark}\textbf{Healthy $W_0$}}\par #1
  \end{tcolorbox}
  \begin{tcolorbox}[enhanced, colback=woneLight, colframe=wone, boxrule=0.5pt,
    arc=2pt, left=3pt, right=3pt, top=1.5pt, bottom=1.5pt,
    fontupper=\scriptsize\raggedright]
    {\color{woneDark}\textbf{Shifted $W_1$}}\par #2
  \end{tcolorbox}
  \end{tcbraster}}
\newcommand{\persisted}[2]{%
  \begin{tcolorbox}[enhanced, colback=woneLight!55, frame hidden, sharp corners,
    borderline west={2pt}{0pt}{wone}, left=5pt, right=3pt, top=2pt, bottom=2pt,
    fontupper=\scriptsize\raggedright, before skip=0pt, after skip=4pt]
    {\color{woneDark}\textbf{#1}\hfill\ding{55}\,stale in $W_1$}\par\vspace{1pt}
    #2
  \end{tcolorbox}}
\newcommand{\recoverable}[1]{%
  {\color{wmaintDark}\ding{51}\,\textbf{Recoverable in $W_1$.}}~#1}
  
\newcommand{\skillpairneg}[3]{%
  #1$\!\rightarrow\!$#2\,%
  \textbf{\textcolor{SkillNegText}{\ensuremath{(#3)}}}%
}

\newcommand{\losscell}[1]{%
  \cellcolor{SkillLoss}\textbf{\ensuremath{#1}}%
}

\newcommand{\improvecell}[1]{%
  \cellcolor{SkillImprove}\textbf{\ensuremath{#1}}%
}

\newcommand{\skillval}[2]{%
  #1\,\textbf{\ensuremath{(#2)}}%
}

\newcommand{\shiftimprovecell}[2]{%
  \cellcolor{SkillImprove}#1\,\textbf{\ensuremath{(#2)}}%
}

\newcommand{\shiftnegcell}[2]{%
  \cellcolor{SkillLoss}#1\,%
  \textbf{\textcolor{SkillNegText}{\ensuremath{(#2)}}}%
}

\newcommand{\maintgaincell}[2]{\shiftimprovecell{#1}{#2}}

\newcommand{\maintnegcell}[2]{\shiftnegcell{#1}{#2}}

\usepackage[T1]{fontenc}

\usepackage[utf8]{inputenc}

\usepackage{microtype}

\usepackage{inconsolata}

\usepackage{graphicx}

\title{When Agent Skills Go Stale:\\Runtime Behavior and Selective Maintenance in Evolving Environments\newline
}

\author{
  \textbf{\"Omer Faruk Akg\"ul}$^{1,2}$,
  \textbf{Shuli Jiang}$^{2}$,
  \textbf{Yuting Zhang}$^{2}$, \\[0.3em]
  \textbf{Yi Zhang}$^{2}$,
  \textbf{Xiaohan Fei}$^{2}$,
  \textbf{Wei Xia}$^{2}$,
  \textbf{Stefano Soatto}$^{2}$ \\[0.3em]
  $^{1}$University of Southern California 
  $^{2}$AWS Agentic AI \\[0.4em]
  {\small\textit{Work done during the first author's internship at AWS.}}
}

\begin{document}
\maketitle

\begin{abstract}
    Agent skills use past experience and current observations to guide decisions and actions on related tasks. This experience is retained in skill state, whose reuse can amortize inference-time reasoning and environment interaction. When environment changes, parts of this state can become outdated while others remain useful. Across four task settings and two models, we compare skill reuse before and after environment changes. Previously beneficial skills can reduce task performance by over 10 percentage points or increase execution cost by nearly 40\% relative to no skill. Task success can also remain unchanged while execution savings decline. These effects depend on which environment-specific details become outdated and the agent’s ability to compensate using current observations. We introduce selective maintenance, which replays checks from successful executions without additional model calls. It verifies whether observed differences conflict with skill state, repairs affected content, and validates updates, rebuilding when local repair is insufficient. No description of the change is supplied. Relative to reusing stale state, maintenance improves task accuracy by up to 67pp and reduces execution steps by up to 37\%. These results suggest that preserving a skill's inference time value as environments evolve requires selectively maintaining its state.
\end{abstract}

\begin{figure*}[!b]
  \centering
  \vspace{-2mm}
  \includegraphics[width=0.99\textwidth]{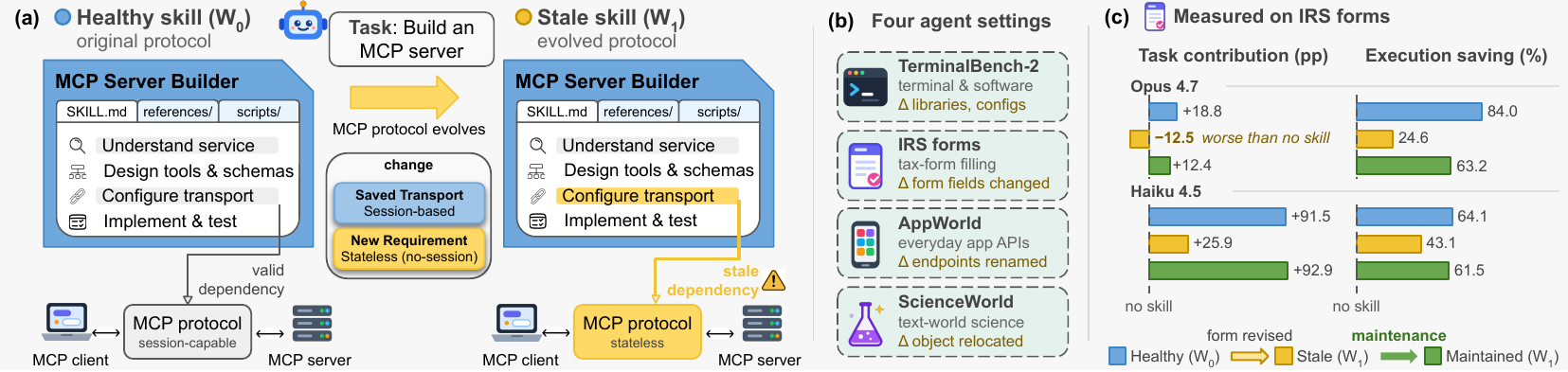}
  \vspace{-2mm}
  \caption{\textbf{Environmental changes can leave skill state partly outdated.}
  (a)~An illustrative change in an MCP server's communication requirements invalidates a saved connection setting while leaving most of the procedure useful.
  (b)~The four evaluation settings and their environment changes.
  (c)~On tax-form completion, benefits over the corresponding no-skill baseline (dashed line) decline after a form revision and recover with maintenance.}
  \label{fig:skill-staleness}
  \vspace{-3mm}
\end{figure*}

\section{Introduction}
\label{sec:intro}

Agent systems increasingly reuse prior experience to guide decisions and actions on related tasks, often through representations referred to as "skills" \citep{wang2023voyageropenendedembodiedagent,wang2024agentworkflowmemory,zheng2025skillweaverwebagentsselfimprove,jiang2026sokagenticskills}. 
Skills use past experience together with current observations to inform future behavior, while experience dependent information retained across executions forms part of a skill's state. Reusing this state allows previously learned decisions and procedures to be reused rather than reconstructed from scratch. In this sense, skills instantiate a broader view of learning as amortizing future inference-time computation: experience retained today can reduce the reasoning and environment interaction required on later tasks \citep{achille2026ai}.

This amortization, however, depends on the continued validity of the retained skill state as the operating environment evolves. A software update, for example, may rename an API endpoint while preserving the rest of a workflow, so most stored knowledge remains useful although one operational choice has become invalid. Figure~\ref{fig:skill-staleness} illustrates the same phenomenon for a Model Context Protocol (MCP) server: a saved connection setting conflicts with a new communication requirement, while the remainder of the procedure remains applicable.

We call skill state \emph{stale} when an environmental change makes some retained information no longer valid for guiding decisions or actions in the current environment. 
Reusing stale state can reduce task performance. Alternatively, the agent may detect the mismatch from current observations and compensate through additional reasoning or environment interaction.
Such runtime recovery can preserve task success without correcting the retained state for future use. 
We therefore measure both task performance and execution cost relative to a \emph{no-skill baseline}, in which the same agent operates without access to the retained skill state.

Prior work evaluates reusable skills \citep{li2026skillsbench,han2026sweskills,liu2026skillswild,lei2026skillevolbench}, studies changing tool interfaces and executable environments \citep{chen2025toolevo,liu2024codeupdatearena,guo2025stabletoolbenchstablelargescalebenchmarking,li2026proevolve}, and examines skill-library evolution \citep{liu2026skillsvote,pu2026skillops,shen2026dynamicskilllifecyclemanagement,fan2026skillguard}.
Our focus is how an environmental change alters the benefits of reusing the same skill state: when it remains useful, when it increases cost or lowers performance, and whether selective updates can recover the benefits of reuse.

We investigate two factors that help explain the effects of stale skill state.
The first is which environment-dependent choices the saved state fixes in advance. Instructions can explain how to find a current field identifier or store an identifier found earlier. We call the extent to which choices are fixed in advance \emph{operational specificity}.
Using the same source material, we construct \emph{procedural} variants that retain inspection and derivation steps, and \emph{cache-heavy} variants that store resolved values and commands without those discovery steps.
The second factor is the agent's ability to recognize and compensate for outdated state. An agent may follow a saved instruction, inspect the current environment, or revise its action using that evidence.

We evaluate two models across four settings: \terminalbench (TB-2) \citep{merrill2026terminalbenchbenchmarkingagentshard}, historical \irs tax forms, \appworld \citep{trivedi2024appworldcontrollableworldapps}, and \scienceworld \citep{wang2022scienceworldagentsmarter5th}.
Each agent combines an LLM with a harness that manages tool use and execution.
For each paired environment change $W_0\!\rightarrow W_1$, the task objective remains fixed. Changes include software and configuration updates, revised form fields, renamed API endpoints, and moved objects.
For each model, we keep the harness fixed and compare reuse of the same saved state against the no-skill baseline in both versions. Execution cost is measured in agent turns or, for \scienceworld, environment steps.

After a change, saved state can still improve performance while providing smaller cost reductions. It can also lower performance despite reducing cost, or perform worse and cost more than the no-skill baseline.
On TB-2 and tax-form completion, cache-heavy variants generally save more work before a change but lose more of their performance benefit afterward. The variants behave similarly on \appworld and \scienceworld, where the change affects choices stored by both.
In the tax-form analysis, the stronger model more reliably overrides outdated instructions after inspecting the current form. Across our tested conditions, however, outdated state sometimes lowers the stronger model's performance below its no-skill baseline, while the weaker model continues to benefit from it.
These findings show why task success alone cannot establish whether reuse still saves work.

The findings also suggest when to maintain skill state.
Checking for any environmental change is too broad, since many changes do not affect the saved content. Waiting for task failure misses additional execution costs, and task outcomes may be unavailable during deployment.
We introduce \emph{selective maintenance}, which constructs checks from saved content and successful executions and records their results in the original environment.
It replays these checks without model calls, compares observations with recorded results, and verifies whether differences conflict with the saved state.
It repairs affected content and validates the result, rebuilding the instructions and code when a conflict cannot be localized or a repair cannot be validated.
The method receives no description of what changed.

Compared with reusing outdated skill state, selective maintenance improves tax-form field accuracy by up to 67 percentage points and reduces \scienceworld environment steps by up to 37\%.
In the tax-form experiments, model tokens saved on subsequent tasks repay the one-time model-token cost of maintenance after nearly 21--42 uses, relative to retiring the outdated state and returning to the no-skill baseline.
Our contributions are:
\begin{itemize}[leftmargin=*,topsep=2pt,itemsep=2pt]
    \item \textbf{A formulation of staleness in skill state.}
    We measure how the performance and execution benefits of the same saved state change across paired environment versions in four settings.

    \item \textbf{An analysis of what makes outdated state useful or harmful.}
    We compare two authoring styles and two models to examine how saved choices and runtime responses affect performance and cost.

    \item \textbf{Selective maintenance of skill state.}
    We use environment checks to identify conflicts, repair affected content, and validate updates, with rebuilding when needed. We evaluate the recovered benefits and the reuse needed to repay maintenance cost.
\end{itemize}

\section{Problem Formulation}
\label{sec:formulation}

\subsection{Agent Execution across Environment Versions}
\label{sec:setting}

We study the state through which a skill carries experience across executions, including reusable instructions, code, and associated resources \citep{jiang2026sokagenticskills,xu2026agentskillslargelanguage}.
Let $\sigma_0$ denote a set of these resources created or validated in an
original environment $W_0$ and reused unchanged after the environment changes
to $W_1$.

Let $\pi$ denote an agent with a fixed base model and execution harness,
and let $x\sim\mathcal{X}$ denote a task drawn from the evaluation
distribution.
We hold the task distribution, objectives, and scoring criteria fixed when
comparing the two environment versions, with and without $\sigma_0$.

Let $s\in\{\varnothing,\sigma_0\}$ indicate whether the agent has access to
$\sigma_0$, with $s=\varnothing$ denoting the \emph{no-skill baseline}.
The resulting trajectory, including observations, actions, and the final
output, is distributed as $\tau\sim P_t^\pi(\cdot\mid x,s)$, where
$t\in\{0,1\}$ indexes environment versions.

\vspace{-1mm}

\subsection{Staleness and Skill Value}
\label{sec:staleness}

Skill state may retain environment-dependent information, such as a file path, identifier, or assumption about API behavior. 
For a given task,  we call $\sigma_0$ \emph{stale} under
$W_0\!\rightarrow W_1$ when information relevant to the task was valid in $W_0$ but no longer holds in $W_1$.
Other parts of the state may remain valid and useful.

\emph{Operational specificity} describes the extent to which skill state resolves environment-dependent details in advance rather than leaving them to be determined during execution.
For example, a skill may store a particular identifier instead of retaining a procedure for recovering the appropriate identifier from the current environment. 
Section~\ref{sec:specificity-capability} studies this distinction using
procedural and cache-heavy variants.

When stale state is reused, the agent may follow outdated information or detect the mismatch from current observations and compensate at runtime. Such recovery can preserve task performance while requiring additional reasoning or environment interaction.
We therefore define \emph{skill value} relative to the same agent without the skill state, with two components: marginal task contribution and execution saving.

Let $R_t(x,\tau)\in[0,1]$ denote the normalized task score under the fixed
scoring criterion, and let $c_t(x,\tau)\geq 0$ denote the cost of executing
the task: agent turns, or environment steps in \scienceworld.
This cost excludes creating or maintaining $\sigma_0$.
Expected task performance and execution cost are
\begin{align}
    U_t^\pi(s)
    &= \mathbb{E}\!\left[R_t(x,\tau)\right], &
    C_t^\pi(s)
    &= \mathbb{E}\!\left[c_t(x,\tau)\right].
    \label{eq:utility-cost}
\end{align}
\vspace{-3mm}

The expectation includes task sampling and trajectory randomness under
$\pi$, $s$, and $W_t$.
We estimate these expectations using mean scores and costs over each
benchmark's evaluation tasks.

Within each environment version, we measure the benefit of reusing
$\sigma_0$ relative to the same agent's no-skill baseline
\citep{han2026sweskills}.
Assuming finite mean costs and $C_t^\pi(\varnothing)>0$, the performance
gain and relative cost reduction are
\begin{align}
    \Delta U_t^\pi
    &=
    U_t^\pi(\sigma_0)-U_t^\pi(\varnothing),\\
    S_t^\pi
    &=
    \frac{
        C_t^\pi(\varnothing)-C_t^\pi(\sigma_0)
    }{
        C_t^\pi(\varnothing)
    },
    \qquad t\in\{0,1\}.
    \label{eq:marginal-skill-value}
\end{align}
Positive $\Delta U_t^\pi$ means higher task performance than the no-skill
baseline.
$S_t^\pi$ expresses the reduction in mean cost as a fraction of mean
baseline cost: positive values indicate lower cost, and negative values
indicate higher cost.
We interpret these quantities jointly, since lower cost can accompany lower
task performance.

We measure changes in these benefits across the environment transition as
\begin{equation}
    L_U^\pi
    =
    \Delta U_0^\pi-\Delta U_1^\pi,
    \qquad
    L_S^\pi
    =
    S_0^\pi-S_1^\pi.
    \label{eq:skill-value-loss}
\end{equation}

\vspace{-1mm}

Positive values indicate a loss of relative benefit; negative values
indicate an improvement.
When $S_t^\pi$ is reported as a percentage, $L_S^\pi$ is a percentage-point
difference.
A change in the no-skill baseline can also change these benefits, so
positive losses alone do not establish staleness.
We call $\Delta U_1^\pi<0$ \emph{negative transfer}: using $\sigma_0$ lowers
task performance below the no-skill baseline in $W_1$.

These quantities characterize the benefit of reusing $\sigma_0$ in
controlled evaluations.
Estimating them requires corresponding no-skill executions, which may be
unavailable when deciding whether to update $\sigma_0$ during deployment.

\vspace{-1mm}
\vspace{-1mm}

\section{Skill Value under Changing Environment}
\label{sec:runtime-results}
\vspace{-2mm}

\begin{table*}[t]
\centering
\caption{
\textbf{Skill value before and after environment changes}
(cache-heavy variants; procedural variants in
Table~\ref{tab:procedural-broad-value}).
Cells report \emph{no-skill $\rightarrow$ $\sigma_0$}, where the same skill state $\sigma_0$ is reused across environment versions; parentheses show marginal task contribution $\Delta U$ or relative execution-cost reduction $S$. Performance is measured by pass rate on \tb, field accuracy on \irs, task success on \appworld, and normalized score on \scienceworld. Execution cost is measured in agent turns, except on \scienceworld, where we use environment steps. 
$L_U$ and $L_S$ measure losses in these relative benefits across the change.
Red marks losses, green improvements, and dark red negative performance
gains in $W_1$. Percentage points are abbreviated as pp.
}
\label{tab:broad-value}

\begingroup
\scriptsize
\setlength{\tabcolsep}{2.15pt}
\renewcommand{\arraystretch}{1.04}

\begin{tabular}{@{}llccc ccc@{}}
\toprule
&
&
\multicolumn{3}{c}{\textbf{Task performance}}
&
\multicolumn{3}{c}{\textbf{Execution cost}}
\\
\cmidrule(lr){3-5}
\cmidrule(lr){6-8}

\textbf{Benchmark}
&
\textbf{Model}
&
\shortstack{\textbf{Original $W_0$}\\[-1pt]
No $\rightarrow$ Skill}
&
\shortstack{\textbf{Changed $W_1$}\\[-1pt]
No $\rightarrow$ Skill}
&
\shortstack{\textbf{Loss}\\[-1pt]$L_U$}
&
\shortstack{\textbf{Original $W_0$}\\[-1pt]
No $\rightarrow$ Skill}
&
\shortstack{\textbf{Changed $W_1$}\\[-1pt]
No $\rightarrow$ Skill}
&
\shortstack{\textbf{Loss}\\[-1pt]$L_S$}
\\
\midrule

\multirow{2}{*}{\tb}
& Opus
& \skillpair{.921}{.968}{+4.7\,\mathrm{pp}}
& \skillpairneg{.895}{.781}{-11.4\,\mathrm{pp}}
& \losscell{16.1\,\mathrm{pp}}
& \skillpair{40.7}{13.6}{66.5\%}
& \skillpair{37.6}{24.2}{35.6\%}
& \losscell{30.8\,\mathrm{pp}}
\\

& Haiku
& \skillpair{.447}{.874}{+42.6\,\mathrm{pp}}
& \skillpair{.447}{.561}{+11.4\,\mathrm{pp}}
& \losscell{31.2\,\mathrm{pp}}
& \skillpair{66.1}{25.9}{60.9\%}
& \skillpair{68.9}{38.9}{43.5\%}
& \losscell{17.4\,\mathrm{pp}}
\\

\addlinespace[1.5pt]

\multirow{2}{*}{\irs}
& Opus
& \skillpair{.772}{.960}{+18.8\,\mathrm{pp}}
& \skillpairneg{.856}{.731}{-12.5\,\mathrm{pp}}
& \losscell{31.3\,\mathrm{pp}}
& \skillpair{10.0}{1.6}{84.0\%}
& \skillpair{10.3}{7.8}{24.6\%}
& \losscell{59.4\,\mathrm{pp}}
\\

& Haiku
& \skillpair{.007}{.922}{+91.5\,\mathrm{pp}}
& \skillpair{.046}{.305}{+25.9\,\mathrm{pp}}
& \losscell{65.6\,\mathrm{pp}}
& \skillpair{13.1}{4.7}{64.1\%}
& \skillpair{12.3}{7.0}{43.1\%}
& \losscell{21.0\,\mathrm{pp}}
\\

\addlinespace[1.5pt]

\multirow{2}{*}{\appworld}
& Opus
& \skillpair{.863}{.961}{+9.8\,\mathrm{pp}}
& \skillpair{.863}{.922}{+5.9\,\mathrm{pp}}
& \losscell{3.9\,\mathrm{pp}}
& \skillpair{20.2}{12.5}{37.9\%}
& \skillpair{22.6}{17.5}{22.2\%}
& \losscell{15.6\,\mathrm{pp}}
\\

& Haiku
& \skillpair{.490}{.549}{+5.9\,\mathrm{pp}}
& \skillpair{.471}{.627}{+15.7\,\mathrm{pp}}
& \improvecell{-9.8\,\mathrm{pp}}
& \skillpair{33.1}{25.5}{22.9\%}
& \skillpair{33.4}{28.7}{13.9\%}
& \losscell{9.0\,\mathrm{pp}}
\\

\addlinespace[1.5pt]

\multirow{2}{*}{\scienceworld}
& Opus
& \skillpair{1.000}{1.000}{+0.0\,\mathrm{pp}}
& \skillpairneg{1.000}{.870}{-13.0\,\mathrm{pp}}
& \losscell{13.0\,\mathrm{pp}}
& \skillpair{17.3}{14.5}{16.2\%}
& \skillpair{19.3}{26.8}{-38.9\%}
& \losscell{55.0\,\mathrm{pp}}
\\

& Haiku
& \skillpair{.859}{1.000}{+14.1\,\mathrm{pp}}
& \skillpair{.927}{1.000}{+7.3\,\mathrm{pp}}
& \losscell{6.8\,\mathrm{pp}}
& \skillpair{52.3}{23.0}{56.1\%}
& \skillpair{55.0}{28.0}{49.0\%}
& \losscell{7.1\,\mathrm{pp}}
\\

\bottomrule
\end{tabular}
\endgroup
\vspace{-3mm}
\end{table*}

We examine how environment changes affect the benefits of reusing skill state.
For each transition, we evaluate Opus~4.7 and Haiku~4.5 with and without
the same skill state in both environments (App.~\ref{app:actors}).
We compare two authoring styles: procedural skills retain how to determine
current environment-dependent choices, while cache-heavy skills retain
more choices resolved in the original environment.
Table~\ref{tab:broad-value} reports cache-heavy results; the procedural
counterpart appears in Table~\ref{tab:procedural-broad-value}.

\vspace{-2mm}

\paragraph{Environment transitions.}
Each transition changes the environment while keeping skill state, the task
objective, and the scoring criterion fixed.
The information needed to adapt remains accessible in $W_1$
(App. Figure~\ref{fig:transition-examples}).
On \tb, we use audited changes to packages, configuration, schemas, and files.
On \irs, we use historical tax-form revisions with renumbered, renamed,
removed, or added fields.
On \appworld, we rename API endpoints and update their documentation.
On \scienceworld, we move the target object in conductivity tasks to the
agent's starting room while leaving the task's location wording unchanged.
Table~\ref{tab:evaluation-populations} gives the details.

\vspace{-1mm}

\subsection{Changes in Performance and Execution Cost}
\label{sec:value-results}

Across the eight comparisons in Table~\ref{tab:broad-value}, execution savings decrease after environment change in all cases, while marginal task contribution decreases in seven. 
Both quantities are measured relative to the no-skill baseline in the same environment.
After the shift, however, the two components need not degrade in the same way.

Opus illustrates three distinct outcomes.
On \appworld, reusing $\sigma_0$ continues to improve task performance and reduce execution cost, but the execution saving is smaller than in the original environment. 
On \irs, it still reduces the number of turns but lowers field accuracy relative to the no-skill baseline.
On \scienceworld, it is harmful on both dimensions, reducing task performance while requiring $38.9\%$ more environment steps than no skill state.
These cases show why skill value must be evaluated along both dimensions: stale state may preserve task benefit while losing execution savings, or reduce execution cost while harming task performance.

\vspace{-1mm}

\subsection{Operational Specificity \& Post-Shift Value}
\label{sec:specificity-capability}

\vspace{-1mm}

\begin{table}[h]
\centering
\caption{
\textbf{Comparison of skill authoring styles.}
Each cell shows \emph{procedural $\rightarrow$ cache-heavy}.
$S_0$ is relative cost reduction in $W_0$ (\%); $L_U$ is loss of
performance gain across the change (pp).
Parentheses give cache-heavy minus procedural differences (pp).
Green highlights greater cost reduction; red highlights greater
performance loss. }
\label{tab:factorial-value}
\begingroup
\scriptsize
\setlength{\tabcolsep}{1.7pt}
\renewcommand{\arraystretch}{1.05}
\begin{tabular}{@{}llcc@{}}
\toprule
\textbf{Benchmark} & \textbf{Model} &
\shortstack{\textbf{Cost reduction $S_0$}\\[-1pt] Proc.\ $\rightarrow$ Cache} &
\shortstack{\textbf{Performance loss $L_U$}\\[-1pt] Proc.\ $\rightarrow$ Cache} \\
\midrule
\multirow{2}{*}{\tb}
& Opus
& \cellcolor{SkillImprove}\skillpair{34.9}{66.5}{+31.5}
& \cellcolor{SkillLoss}\skillpair{$-$6.8}{16.1}{+22.9} \\
& Haiku
& \cellcolor{SkillImprove}\skillpair{29.7}{60.9}{+31.2}
& \skillpair{13.8}{31.2}{+17.5} \\
\addlinespace[1.5pt]
\multirow{2}{*}{\irs}
& Opus
& \cellcolor{SkillImprove}\skillpair{65.0}{84.0}{+19.0}
& \cellcolor{SkillLoss}\skillpair{12.6}{31.3}{+18.7} \\
& Haiku
& \skillpair{65.0}{64.1}{-0.9}
& \cellcolor{SkillLoss}\skillpair{48.8}{65.6}{+16.8} \\
\addlinespace[1.5pt]
\multirow{2}{*}{\appworld}
& Opus  & \skillpair{33.8}{37.9}{+4.1}  & \skillpair{3.9}{3.9}{+0.0} \\
& Haiku & \skillpair{32.7}{22.9}{-9.8}  & \skillpair{$-$7.8}{$-$9.8}{-2.0} \\
\addlinespace[1.5pt]
\multirow{2}{*}{\scienceworld}
& Opus  & \skillpair{11.6}{16.2}{+4.6}  & \skillpair{14.1}{13.0}{-1.1} \\
& Haiku & \skillpair{66.1}{56.1}{-10.0} & \skillpair{6.8}{6.8}{+0.0} \\
\bottomrule
\end{tabular}
\endgroup
\vspace{-3mm}
\end{table}

\vspace{-1mm}

We study operational specificity through the two authoring styles,
using the same author model and the same $W_0$ material for both variants
(App.~\ref{app:skill-construction}).
The cache-heavy authoring prompt excludes discovery steps and fallbacks,
so the variants differ in both the choices they retain and the procedures
they provide for finding current information.

\textbf{Procedural lookup helps recover changed paths and field names.}
The \tb and \irs examples illustrate this behavior
(App. Figure~\ref{fig:transition-examples}).
Across these two settings and two models, cache-heavy variants lose more
of their performance gain after the change in all four comparisons.
They also provide greater cost reduction in $W_0$ in three of the four
comparisons (Table~\ref{tab:factorial-value}).

The \irs results with Opus illustrate the tradeoff.
Both variants initially improve field accuracy by similar amounts.
After the form changes, the procedural variant retains a 6.2-point gain
over no skill, while the cache-heavy variant lowers accuracy by
12.5 points relative to the same baseline.
The procedural variant loses almost all of its original cost reduction,
while the cache-heavy still saves turns.

The pattern is not universal. On \appworld and \scienceworld, the environment change invalidates information retained by both variants, such as an API endpoint name or a task-specified location. Their performance losses are therefore similar, and neither variant consistently yields greater execution savings in $W_0$. 
Procedural lookup steps help recover changed paths and field names in the
\tb and \irs examples (App. Figure~\ref{fig:transition-examples}), but
provide little advantage when both variants retain the same outdated
endpoint or location choice.

\subsection{Realized Skill Value Depends on the Agent}
\label{sec:actor-capability}

\textbf{The same stale skill state can help one agent and hurt another.}
With the cache-heavy \irs skill state held fixed, Opus achieves higher
absolute field accuracy than Haiku ($73.1\%$ versus $30.5\%$).
Relative to each agent's own no-skill baseline, however, reuse lowers
Opus's accuracy while improving Haiku's.
Higher absolute accuracy therefore does not imply a greater benefit
from reuse.

This pattern extends beyond \irs.
Across all four benchmarks, Haiku retains positive performance gains
with both authoring styles, while Opus sometimes performs better without
the skill (Tables~\ref{tab:broad-value} and~\ref{tab:procedural-broad-value}).
Thus, realized skill value depends not only on what information is retained in the state, but also on how the agent interprets, verifies, and compensates for that information at runtime. We next examine executions on the revised \irs forms to characterize these different responses to stale state.

\vspace{-1mm}

\subsection{How Agents Respond to Stale Skill State}
\label{sec:runtime-response}
\vspace{-1mm}

\begin{table}[h]
\centering
\caption{
\textbf{Responses to stale skill state on \irs.}
Each row: 50 trajectories (5 forms $\times$ 10 applicants).
\emph{Inspects}: reads the current form before submitting the first
final program.
\emph{Obsolete write}: writes to a field present in $W_0$ but absent
from $W_1$; counted among inspecting trajectories.
$\Delta$: inspecting minus non-inspecting differences in field
accuracy (pp) and total turns, within forms containing both groups
(Appendix~\ref{app:runtime-annotations}).
Red highlights lower inspection rates, more obsolete writes, or
additional turns; green, larger accuracy differences.}
\label{tab:runtime-main}
\scriptsize
\setlength{\tabcolsep}{3pt}
\renewcommand{\arraystretch}{1.05}
\begin{tabular}{@{}llcccc@{}}
\toprule
\textbf{Model} & \textbf{Style} & \textbf{Inspects} &
\shortstack{\textbf{Obsolete write}\\after insp.} &
\shortstack{$\Delta$\textbf{Acc.}\\(pp)} & $\Delta$\textbf{Turns} \\
\midrule
Opus  & Proc.  & 43/50
      & 5/43
      & \cellcolor{SkillImprove}$+24$
      & \cellcolor{SkillLoss}$+6.2$ \\
Opus  & Cache  & \cellcolor{SkillLoss}30/50
      & 2/30
      & \cellcolor{SkillImprove}$+68$
      & \cellcolor{SkillLoss}$+10.4$ \\
\addlinespace[1.5pt]
Haiku & Proc.  & 41/50
      & \cellcolor{SkillLoss}14/41
      & $+9$
      & \cellcolor{SkillLoss}$+7.1$ \\
Haiku & Cache  & 39/50
      & \cellcolor{SkillLoss}22/39
      & $+3$
      & \cellcolor{SkillLoss}$+6.5$ \\
\bottomrule
\end{tabular}
\vspace{-3mm}
\end{table}

We examine the \irs trajectories under both authoring styles, recording
whether the agent inspects the current form and whether it subsequently
writes to an obsolete field name (Table~\ref{tab:runtime-main}).
Obsolete writes indicate use of outdated identifiers; they also occur
in some no-skill runs (App.~\ref{app:runtime-annotations}).

\textbf{Reading current information does not ensure that the agent uses it.}
Both agents frequently inspect the current form, but their subsequent
actions differ.
Across the two styles, Opus writes to obsolete fields in 7 of 73 inspecting
trajectories, compared with 36 of 80 for Haiku.
Within-form comparisons associate inspection with higher field accuracy
and more turns, with larger accuracy differences for Opus in both styles.
These comparisons describe complete trajectories and do not isolate
the causal effect of inspection.

The cache-heavy Opus runs illustrate how fewer turns can accompany worse
task performance.
On average, the non-inspecting group performs worse than the corresponding
forms' no-skill baselines and accounts for the aggregate turn reduction.
The inspecting group improves accuracy but uses more turns than those
baselines (App.~\ref{app:runtime-annotations}).
The benefit of reusing stale state thus depends on whether the agent
checks current information and how it uses that information afterward.
Section~\ref{sec:maintenance} examines whether updating the state can
preserve the benefits of reuse for future tasks.

\section{Selective Skill Maintenance}
\label{sec:maintenance}

\begin{figure*}[t]
    \centering
    \includegraphics[width=\textwidth]{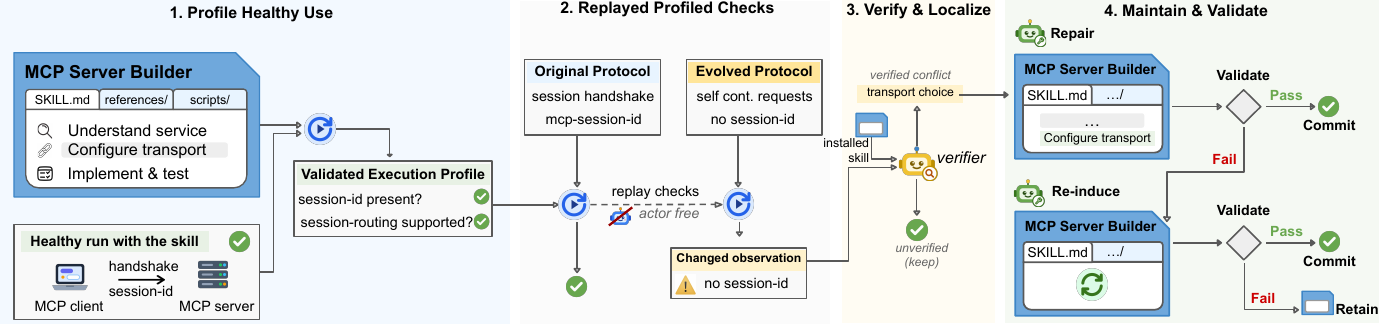}
    \caption{
    \textbf{Selective skill maintenance.}
    Checks derived automatically from retained skill state and successful executions are stored with reference observations and source links, then replayed before reuse without model calls. Verification determines whether observed changes conflict with the retained state and localizes affected content. An LLM selectively repairs that state, rebuilding when repair or localization is insufficient. Candidate updates are validated for task performance and execution cost before replacing the existing state.
    }
    \label{fig:maintenance}
    \vspace{-3mm}
\end{figure*}

An agent may compensate for stale skill state by inspecting the
environment and correcting its actions.
If the retained state remains unchanged, however, future executions may have to repeat the same corrective work. 
Updating only the affected state allows later tasks to reuse the correction while preserving information that remains valid.
We therefore introduce \emph{selective maintenance}, which detects conflicts between retained skill state and the current environment, repairs the affected state, and validates the update 
(Figure~\ref{fig:maintenance}).
The task-performing agent remains fixed; maintenance modifies only the
reusable skill state.

\subsection{Detecting Conflicts \& Updating Skill State}
\label{sec:maintenance-method}

\paragraph{Constructing execution-grounded checks.}
We construct checks from retained skill state $\sigma_0$ and from operations that succeed when the skill is executed in $W_0$.
Rule-based extraction identifies named paths, commands, and identifiers
in its instructions and code.
To check operations such as field writes, API calls, and environment
actions, we execute skill routines or replay code from successful runs.
The particular paths and operations to check are selected automatically,
using observation routines shared across skills.
For the supported operations, construction makes no model calls and
requires no individually written tests for each skill
(Appendix~\ref{app:lifecycle}).

Detecting a skill-relevant change requires both a prior observation for comparison and a link back to the state that depends on it. 
For example, if the retained state refers to a script at a particular path, the profile may record a path-existence check, its Boolean result in $W_0$, and a reference to the retained path. If the file later disappears, the changed observation identifies a potential conflict and the reference localizes the affected state.

The resulting \emph{execution profile} is
\begin{equation}
    M_0
    =
    \left\{
        \bigl(p_j,q_j,e_j^{(0)}\bigr)
    \right\}_{j=1}^{J},
    \label{eq:execution-profile}
\end{equation}
where $q_j$ is an executable check, $e_j^{(0)}=q_j(W_0)$ is its recorded
output, and $p_j$ identifies 
the corresponding element of skill state or validated execution.
To make later comparisons meaningful, we repeat checks in $W_0$ and
discard those whose outputs vary across runs.
The recorded observations capture environment properties relevant to skill execution rather than task answers.
For example, form checks use synthetic inputs with distinct placeholder
values to record where each write lands, excluding applicant data.
Appendix~\ref{app:lifecycle} gives recording and filtering details.

\paragraph{Checking before reuse.}
Before reuse, the maintainer runs the checks in the current environment
$W_t$ without model calls.
Comparing their outputs with the reference observations from $W_0$ yields
\[
    D_t=\{\,j:q_j(W_t)\neq e_j^{(0)}\,\}.
\]
If $D_t$ is empty, $\sigma_0$ is reused unchanged.
Otherwise, the detected differences are passed to verification to determine whether they conflict with the retained state.

\paragraph{Verifying differences.}
A changed observation does not necessarily imply stale skill state. 
Verification checks whether each difference conflicts with information on which $\sigma_0$ relies and localizes the affected state. 
Simple conflicts are resolved directly from the execution profile, and when interpretation is needed, an LLM verifier compares the relevant skill state with the original and current observations 
(Appendix~\ref{app:lifecycle}).
No external changelog is needed, and unsupported differences leave $\sigma_0$ unchanged.

\paragraph{Repairing and rebuilding.}
For a localized conflict, an LLM maintainer updates only the affected skill state while preserving the remainder.
The candidate may be refined using validation feedback
\citep{madaan2023selfrefineiterativerefinementselffeedback,chen2023teachinglargelanguagemodels}.
If the conflict cannot be localized or repair repeatedly fails validation, the maintainer rebuilds the skill state in the current environment using the original
authoring procedure, without access to the stale state.
Repair is attempted first to preserve still-valid information and avoid unnecessary reconstruction.

\paragraph{Validating updates.}
We execute each candidate on validation tasks in the current environment
and measure both task performance and execution cost.
Validation assesses whether an update recovers task performance
and execution savings.
Candidates are accepted according to performance thresholds
and cost targets, with primary and fallback criteria specified
in Appendix~\ref{app:lifecycle}.
If no candidate is accepted, the existing state is retained.
Validation task instances or sampled runs are disjoint from reported evaluation;
held-out runs, answers, and reference conflict labels are available only
to the evaluator (Appendix Table~\ref{tab:information-boundary}).

\subsection{Effects of Selective Maintenance}
\label{sec:maintenance-restoration}

\begin{table}[t]
\centering
\caption{
\textbf{Task performance and execution cost after maintenance (Haiku~4.5).}
No maintenance (No maint.) reuses stale skill state unchanged.
Selective follows the repair-first rule.
Cells report performance or cost; parentheses give performance gains
(pp) or cost reductions relative to the no-skill baseline in $W_1$.
Green marks better means than No maintenance; shading does not indicate
statistical significance.
Cost is agent turns, or environment steps on \scienceworld.
Opus and single-operator results appear in
Appendix Table~\ref{tab:maint-policies}.}
\label{tab:maintenance-results}

\begingroup
\scriptsize
\setlength{\tabcolsep}{1.8pt}
\renewcommand{\arraystretch}{1.05}

\resizebox{\columnwidth}{!}{%
\begin{tabular}{@{}llcccc@{}}
\toprule
\textbf{Benchmark}
&
\textbf{Skill}
&
\shortstack{\textbf{No maint.}\\\textbf{Task}}
&
\shortstack{\textbf{Selective}\\\textbf{Task}}
&
\shortstack{\textbf{No maint.}\\\textbf{Cost}}
&
\shortstack{\textbf{Selective}\\\textbf{Cost}}
\\
\midrule

\multirow{2}{*}{\tb}
& Cache
& \skillval{.561}{+11.4\,\mathrm{pp}}
& \maintgaincell{\textbf{.649}}{+20.2\,\mathrm{pp}}
& \skillval{38.9}{43.5\%}
& \maintgaincell{\textbf{32.1}}{53.4\%}
\\

& Proc.
& \skillval{.711}{+26.3\,\mathrm{pp}}
& \maintgaincell{\textbf{.798}}{+35.1\,\mathrm{pp}}
& \skillval{60.5}{12.1\%}
& \maintgaincell{\textbf{59.7}}{13.3\%}
\\

\midrule

\multirow{2}{*}{\irs}
& Cache
& \skillval{.305}{+25.9\,\mathrm{pp}}
& \maintgaincell{\textbf{.975}}{+92.9\,\mathrm{pp}}
& \skillval{7.0}{43.1\%}
& \maintgaincell{\textbf{4.7}}{61.5\%}
\\

& Proc.
& \skillval{.482}{+43.6\,\mathrm{pp}}
& \maintgaincell{\textbf{.799}}{+75.3\,\mathrm{pp}}
& \skillval{7.2}{41.6\%}
& \maintgaincell{\textbf{6.5}}{46.7\%}
\\

\midrule

\multirow{2}{*}{\appworld}
& Cache
& \skillval{.627}{+15.7\,\mathrm{pp}}
& \maintgaincell{\textbf{.667}}{+19.6\,\mathrm{pp}}
& \skillval{28.7}{13.9\%}
& \maintgaincell{\textbf{23.3}}{30.3\%}
\\

& Proc.
& \skillval{.627}{+15.7\,\mathrm{pp}}
& \skillval{\textbf{.569}}{+9.8\,\mathrm{pp}}
& \skillval{25.7}{22.9\%}
& \maintgaincell{\textbf{24.8}}{25.8\%}
\\

\midrule

\multirow{2}{*}{\scienceworld}
& Cache
& \skillval{1.000}{+7.3\,\mathrm{pp}}
& \skillval{\textbf{1.000}}{+7.3\,\mathrm{pp}}
& \skillval{28.0}{49.0\%}
& \maintgaincell{\textbf{18.6}}{66.1\%}
\\

& Proc.
& \skillval{1.000}{+7.3\,\mathrm{pp}}
& \skillval{\textbf{1.000}}{+7.3\,\mathrm{pp}}
& \skillval{25.6}{53.4\%}
& \maintgaincell{\textbf{16.1}}{70.7\%}
\\

\bottomrule
\end{tabular}%
}

\endgroup
\vspace{-3mm}
\end{table}

We compare \emph{Selective} maintenance with \emph{No maintenance}, which reuses skill state unchanged while allowing the agent to inspect the current environment and compensate at runtime.
Table~\ref{tab:maintenance-results} reports Haiku results; Opus results
appear in Appendix Table~\ref{tab:maint-policies}.
Repair-only and rebuild-only policies use the same verified conflicts
and validation rules.
We evaluate one maintenance event before post-change reuse.\footnote{
Automatic profile refresh after an accepted update is left to future
work; the experiments do not evaluate successive maintenance events.}

\paragraph{Restoring performance and efficiency.}
Compared with No maintenance, Selective improves mean task performance
on \tb{} and \irs, with a gain of up to 67 percentage points in tax-form
field accuracy (Table~\ref{tab:maintenance-results}).
On \scienceworld, the stale state already achieves the maximum task
score; maintenance preserves this performance while reducing environment
steps by up to 37\%.
These outcomes reflect the two reasons to maintain skill state:
recovering performance when stale state causes errors, and
reducing the work needed to complete tasks correctly.

Mean execution cost decreases in all eight settings, although the reductions
are small for procedural variants on \tb{} and \appworld.
The procedural \appworld{} variant also has lower mean task performance
after maintenance; paired estimates are reported in
Appendix Table~\ref{tab:maint-paired}.

\paragraph{Repair vs. rebuilding.}
Most maintenance cases end with an accepted local repair.
Rebuilding produces an accepted update in some cases where repair fails
validation, including the procedural \scienceworld{} skill.
Neither operation consistently yields the more accurate or less costly
skill.
These results support keeping rebuilding available when a local update
is insufficient (Appendix~\ref{app:maintenance-protocol}).
The cost advantage of trying repair first is examined below.

\paragraph{Detecting relevant changes.}
We additionally evaluate detection on environment changes that invalidate retained skill state and on controls where the state remains valid. 
Verification rejects most replay flags in the \tb{} controls, showing
why a changed observation needs further assessment.
Some controls still lead to unnecessary repairs, and replay misses
conflicts caused by changes to file contents outside the properties
recorded by its checks.
Appendix~\ref{app:monitoring-analysis} reports the control designs and
counts at each stage.

\subsection{Cost and Amortization}
\label{sec:maintenance-cost}

Because checks run before reuse, their cost matters even when no update
is needed.
Replay makes no model calls; measured median runtimes range from 0.2 to
4.3 seconds on \irs, \appworld, and \scienceworld.
Appendix Table~\ref{tab:replay-timing} reports the timings and measurement
setup; \tb{} replay was not timed.

When an update is needed, local repair can reduce the work required to
produce it.
On \irs, rebuilding consumes 1.24--2.28$\times$ as many model tokens
as repair, counting verification, candidate generation, and acceptance
validation (Appendix~\ref{app:cost-accounting}).
This supports trying repair before paying for a complete rebuild.

\begin{table}[t]
\centering
\caption{
\textbf{Model-token cost and estimated payback on \irs.}
$K_{\mathrm{sel}}$ counts verification, candidate generation, and
acceptance validation for a maintenance event.
$G_{\varnothing}$ and $G_{\mathrm{sel}}$ are model tokens per subsequent
no-skill and Selective execution.
$n^\star_{\mathrm{retire}}$ estimates the future uses needed to repay
maintenance relative to retiring the stale skill and using no skill.
Values macro-average over five form families.
}
\label{tab:maintenance-amortization}

\begingroup
\scriptsize
\setlength{\tabcolsep}{1.7pt}
\renewcommand{\arraystretch}{1.04}

\resizebox{\columnwidth}{!}{%
\begin{tabular}{@{}llrrrrr@{}}
\toprule
\textbf{Model}
&
\textbf{Skill}
&
\shortstack{\textbf{Maint.}\\$K_{\mathrm{sel}}$}
&
\shortstack{\textbf{No skill}\\$G_{\varnothing}$}
&
\shortstack{\textbf{Selective}\\$G_{\mathrm{sel}}$}
&
\shortstack{\textbf{Save/use}\\$\Delta G$}
&
\shortstack{\textbf{Payback}\\$n^\star_{\mathrm{retire}}$}
\\
\midrule

Haiku
& Proc.
& 151.1K
& 14.58K
& 10.98K
& 3.60K
& \textbf{41.9}
\\

Haiku
& Cache
& 142.8K
& 14.58K
& 7.85K
& 6.74K
& \textbf{21.2}
\\

Opus
& Proc.
& 122.6K
& 5.54K
& 1.87K
& 3.67K
& \textbf{33.4}
\\

Opus
& Cache
& 91.1K
& 5.54K
& 1.97K
& 3.57K
& \textbf{25.5}
\\

\bottomrule
\end{tabular}%
}

\endgroup
\vspace{-3mm}
\end{table}

We next ask how many future uses would recover the model tokens spent
on maintenance.
One alternative is to retire the stale skill
\citep{zhang2026ratchetreliablellmjudge} and return to the no-skill
baseline.
This avoids the update cost but may require the agent to repeat work
previously captured by the skill.
Let $K_{\mathrm{sel}}$ denote the model tokens spent on verification,
candidate generation, and acceptance validation, excluding held-out
evaluation.
Let $G_{\varnothing}$ and $G_{\mathrm{sel}}$ denote mean model tokens per
subsequent task without a skill and after Selective maintenance.
When $G_{\varnothing}>G_{\mathrm{sel}}$, the estimated payback relative
to retirement is
\begin{equation}
    n^\star_{\mathrm{retire}}
    =
    \frac{K_{\mathrm{sel}}}
         {G_{\varnothing}-G_{\mathrm{sel}}}.
    \label{eq:token-amortization-main}
\end{equation}

The averages in Table~\ref{tab:maintenance-amortization} give an estimated
21.2--41.9 uses to repay maintenance.
This assumes future tasks have the measured average token costs and
require no further maintenance during that period.
The estimate concerns model tokens on \irs; environment computation
and latency are outside this accounting.

Reusing stale state provides a different baseline. 
Stale execution may already use few tokens while producing incorrect outputs, so maintenance need not reduce token cost relative to No maintenance even when it substantially improves task performance.
Appendix~\ref{app:cost-accounting} reports payback relative to
No maintenance.
Thus, maintenance should be viewed as an amortization decision over both retained task value and future execution cost: its one-time cost is justified when repeated reuse recovers benefits that would otherwise be lost through retirement or stale execution.
\section{Related Work}
\label{sec:related}

\textbf{Agent skills.}
Agents induce reusable skills and workflows from experience
\citep{wang2023voyageropenendedembodiedagent,wang2024agentworkflowmemory,zheng2025skillweaverwebagentsselfimprove,wang2025inducingprogrammaticskillsagentic},
and evaluations show their gains depend on writing, retrieval, and task fit
\citep{li2026skillsbench,han2026sweskills,liu2026skillswild,lei2026skillevolbench},
and SWE-Skills-Bench traces some harms to version-mismatched guidance.
We instead track the same skill's task and execution value across a
$W_0\!\rightarrow\!W_1$ change. \textbf{Changing environments.}
Prior work tests how agents cope with evolving tools, APIs, and environments
\citep{chen2025toolevo,liu2024codeupdatearena,guo2025stabletoolbenchstablelargescalebenchmarking,li2026proevolve};
we measure what such change does to a persistent skill's value. \textbf{Skill lifecycle and maintenance.}
Recent work retains, repairs, or retires library skills
\citep{liu2026skillsvote,pu2026skillops,shen2026dynamicskilllifecyclemanagement,zhang2026ratchetreliablellmjudge,liu2026skillforgeforgingdomainspecificselfevolving};
Ratchet shows unmaintained libraries can fall below no skill.
Most closely related, SkillGuard checks contracts from skill
documentation against the live environment to localize repair
\citep{fan2026skillguard}.
Our checks are instead grounded in validated skill use, and we evaluate
maintenance by the task and execution value it restores and its token cost.
\section{Conclusion}
\label{sec:conclusion}

Skills retain experience in reusable state that can amortize reasoning and environment interaction across tasks. As environments evolve, some of this state may become stale while other information remains useful. Agents may compensate at runtime, but such recovery can increase execution cost without updating the state. Selective maintenance detects conflicts from prior successful executions and updates only affected state. Our results show that this can restore task performance and reduce execution cost, preserving the value of past experience as environments change.

\clearpage

\section*{Limitations}

Our study covers four datasets and two LLMs from one model family, and
uses controlled transitions that each change one thing a skill could have
persisted, so that every loss can be traced to a specific stale commitment.
Deployed environments may change several things at once and over longer
horizons, and other actors may respond to stale content differently;
extending the analysis to such settings is a natural next step.
The maintenance lifecycle is shared across substrates, but each new
environment may need an adapter that defines what a check observes.

\section*{Use of AI Assistants}
We used LLM agents to construct skills with past experience, execute tasks, and perform the agentic maintenance steps described in the paper. AI assistants also assisted with implementing parts of the experimental code, summarizing material, and revising text for clarity and readability. The authors directed the research and writing, made the final methodological and editorial decisions, and take responsibility for the code, analyses, and manuscript.


\bibliography{custom}

@misc{chen2025toolevo,
      title={Learning Evolving Tools for Large Language Models}, 
      author={Guoxin Chen and Zhong Zhang and Xin Cong and Fangda Guo and Yesai Wu and Yankai Lin and Wenzheng Feng and Yasheng Wang},
      year={2025},
      eprint={2410.06617},
      archivePrefix={arXiv},
      primaryClass={cs.CL},
      url={https://arxiv.org/abs/2410.06617}, 
}

@misc{liu2024codeupdatearena,
      title={CodeUpdateArena: Benchmarking Knowledge Editing on API Updates}, 
      author={Zeyu Leo Liu and Shrey Pandit and Xi Ye and Eunsol Choi and Greg Durrett},
      year={2025},
      eprint={2407.06249},
      archivePrefix={arXiv},
      primaryClass={cs.CL},
      url={https://arxiv.org/abs/2407.06249}, 
}

@misc{wang2024agentworkflowmemory,
      title={Agent Workflow Memory}, 
      author={Zora Zhiruo Wang and Jiayuan Mao and Daniel Fried and Graham Neubig},
      year={2024},
      eprint={2409.07429},
      archivePrefix={arXiv},
      primaryClass={cs.CL},
      url={https://arxiv.org/abs/2409.07429}, 
}

@misc{zheng2025skillweaverwebagentsselfimprove,
      title={SkillWeaver: Web Agents can Self-Improve by Discovering and Honing Skills}, 
      author={Boyuan Zheng and Michael Y. Fatemi and Xiaolong Jin and Zora Zhiruo Wang and Apurva Gandhi and Yueqi Song and Yu Gu and Jayanth Srinivasa and Gaowen Liu and Graham Neubig and Yu Su},
      year={2025},
      eprint={2504.07079},
      archivePrefix={arXiv},
      primaryClass={cs.AI},
      url={https://arxiv.org/abs/2504.07079}, 
}

@misc{han2026sweskills,
      title={SWE-Skills-Bench: Do Agent Skills Actually Help in Real-World Software Engineering?}, 
      author={Tingxu Han and Yi Zhang and Wei Song and Chunrong Fang and Zhenyu Chen and Youcheng Sun and Lijie Hu},
      year={2026},
      eprint={2603.15401},
      archivePrefix={arXiv},
      primaryClass={cs.SE},
      url={https://arxiv.org/abs/2603.15401}, 
}

@misc{pu2026skillops,
      title={SkillOps: Managing LLM Agent Skill Libraries as Self-Maintaining Software Ecosystems}, 
      author={Hongji Pu and Xinyuan Song and Liang Zhao},
      year={2026},
      eprint={2605.13716},
      archivePrefix={arXiv},
      primaryClass={cs.SE},
      url={https://arxiv.org/abs/2605.13716}, 
}

@misc{shen2026dynamicskilllifecyclemanagement,
      title={Dynamic Skill Lifecycle Management for Agentic Reinforcement Learning}, 
      author={Junhao Shen and Teng Zhang and Xiaoyan Zhao and Hong Cheng},
      year={2026},
      eprint={2605.10923},
      archivePrefix={arXiv},
      primaryClass={cs.LG},
      url={https://arxiv.org/abs/2605.10923}, 
}

@misc{zhang2026ratchetreliablellmjudge,
      title={Ratchet: How Reliable Must an LLM Judge Be to Retire a Skill?}, 
      author={Xing Zhang and Yanwei Cui and Guanghui Wang and Ziyuan Li and Wei Qiu and Bing Zhu and Peiyang He},
      year={2026},
      eprint={2605.22148},
      archivePrefix={arXiv},
      primaryClass={cs.AI},
      url={https://arxiv.org/abs/2605.22148}, 
}

@misc{li2026proevolve,
      title={The World Won't Stay Still: Programmable Evolution for Agent Benchmarks}, 
      author={Guangrui Li and Yaochen Xie and Yi Liu and Ziwei Dong and Xingyuan Pan and Tianqi Zheng and Jason Choi and Michael J. Morais and Binit Jha and Shaunak Mishra and Bingrou Zhou and Chen Luo and Monica Xiao Cheng and Dawn Song},
      year={2026},
      eprint={2603.05910},
      archivePrefix={arXiv},
      primaryClass={cs.AI},
      url={https://arxiv.org/abs/2603.05910}, 
}

@misc{wang2023voyageropenendedembodiedagent,
      title={Voyager: An Open-Ended Embodied Agent with Large Language Models}, 
      author={Guanzhi Wang and Yuqi Xie and Yunfan Jiang and Ajay Mandlekar and Chaowei Xiao and Yuke Zhu and Linxi Fan and Anima Anandkumar},
      year={2023},
      eprint={2305.16291},
      archivePrefix={arXiv},
      primaryClass={cs.AI},
      url={https://arxiv.org/abs/2305.16291}, 
}

@misc{li2026skillsbench,
      title={SkillsBench: Benchmarking How Well Agent Skills Work Across Diverse Tasks}, 
      author={Xiangyi Li and Yimin Liu and Wenbo Chen and Bingran You and Zonglin Di and Yifeng He and Shenghan Zheng and Kyoung Whan Choe and Jiankai Sun and Shuyi Wang and Chujun Tao and Binxu Li and Xuandong Zhao and Hejia Geng and Xiaojun Wu and Junwei Zhou and Xiaokun Chen and Hanwen Xing and Yubo Li and Qunhong Zeng and Di Wang and Yuanli Wang and Roey Ben Chaim and Penghao Jiang and Haotian Shen and Luyang Kong and Xinyi Liu and Runhui Wang and Xuanqing Liu and Jiachen Li and Xin Lan and Yueqian Lin and Wengao Ye and Junwei He and Songlin Li and Yue Zhang and Yipeng Gao and Yijiang Li and Ze Ma and Liqiang Jing and Tianyu Wang and Kaixin Li and Yiqi Xue and Haoran Lyu and Yizhuo He and Yuchen Tian and Shutong Wu and Bowei Wang and Yixuan Gao and Bo Chen and Litong Liu and Sikai Cheng and Jiajun Bao and Shuaicheng Tong and Shuwen Xu and Terry Yue Zhuo and Tinghan Ye and Qi Qi and Miao Li and Longtai Liao and Zelin Tan and Chang Shi and Xilin Tang and Srinath Tankasala and Boqin Yuan and Yaoyao Qian and Jianhong Tu and Chenguang Wang and Yizhou Sun and Wei Wang and Aaron Taylor and Ziyue Yang and Changkun Guan and Zhikang Dong and Xinyu Zhang and Steven Dillmann and Han-chung Lee and Dawn Song},
      year={2026},
      eprint={2602.12670},
      archivePrefix={arXiv},
      primaryClass={cs.AI},
      url={https://arxiv.org/abs/2602.12670}, 
}

@article{liu2026skillswild,
  title   = {How Well Do Agentic Skills Work in the Wild:
             Benchmarking LLM Skill Usage in Realistic Settings},
  author  = {Liu, Yujian and Ji, Jiabao and An, Li and
             Jaakkola, Tommi and Zhang, Yang and Chang, Shiyu},
  journal = {arXiv preprint arXiv:2604.04323},
  year    = {2026}
}

@article{lei2026skillevolbench,
  title   = {{SkillEvolBench}: Benchmarking the Evolution from Episodic Experience
             to Procedural Skills},
  author  = {Lei, Yingtie and Wan, Zhongwei and Zhang, Jiankun and
             Alam, Samiul and Zhong, Zixuan and Huang, Peizhou and
             Wang, Xin and Zhang, Jingxuan and Zhou, Donghao and
             Hsieh, Yunta and Dou, Zhihao and Shen, Hui and Xu, Yan and
             Dimitriadis, Dimitrios and Zhang, Tuo and Zhang, Mi},
  journal = {arXiv preprint arXiv:2605.24117},
  year    = {2026}
}

@article{liu2026skillsvote,
  title   = {{SkillsVote}: Lifecycle Governance of Agent Skills from Collection,
             Recommendation to Evolution},
  author  = {Liu, Hongyi and Yang, Haoyan and Jiang, Tao and Tang, Bo and
             Xiong, Feiyu and Luo, Yuyu and Li, Zhiyu},
  journal = {arXiv preprint arXiv:2605.18401},
  year    = {2026}
}

@article{fan2026skillguard,
  title   = {Skill Drift Is Contract Violation:
             Proactive Maintenance for LLM Agent Skill Libraries},
  author  = {Fan, Linfeng and Tian, Yuan and Li, Ziwei and Lu, Zhiwu},
  journal = {arXiv preprint arXiv:2605.10990},
  year    = {2026}
}

@article{achille2026ai,
  title   = {AI Agents as Universal Task Solvers},
  author  = {Achille, Alessandro and Soatto, Stefano},
  journal = {Entropy},
  volume  = {28},
  number  = {3},
  pages   = {332},
  year    = {2026},
  doi     = {10.3390/e28030332}
}

@misc{merrill2026terminalbenchbenchmarkingagentshard,
      title={Terminal-Bench: Benchmarking Agents on Hard, Realistic Tasks in Command Line Interfaces}, 
      author={Mike A. Merrill and Alexander G. Shaw and Nicholas Carlini and Boxuan Li and Harsh Raj and Ivan Bercovich and Lin Shi and Jeong Yeon Shin and Thomas Walshe and E. Kelly Buchanan and Junhong Shen and Guanghao Ye and Haowei Lin and Jason Poulos and Maoyu Wang and Marianna Nezhurina and Jenia Jitsev and Di Lu and Orfeas Menis Mastromichalakis and Zhiwei Xu and Zizhao Chen and Yue Liu and Robert Zhang and Leon Liangyu Chen and Anurag Kashyap and Jan-Lucas Uslu and Jeffrey Li and Jianbo Wu and Minghao Yan and Song Bian and Vedang Sharma and Ke Sun and Steven Dillmann and Akshay Anand and Andrew Lanpouthakoun and Bardia Koopah and Changran Hu and Etash Guha and Gabriel H. S. Dreiman and Jiacheng Zhu and Karl Krauth and Li Zhong and Niklas Muennighoff and Robert Amanfu and Shangyin Tan and Shreyas Pimpalgaonkar and Tushar Aggarwal and Xiangning Lin and Xin Lan and Xuandong Zhao and Yiqing Liang and Yuanli Wang and Zilong Wang and Changzhi Zhou and David Heineman and Hange Liu and Harsh Trivedi and John Yang and Junhong Lin and Manish Shetty and Michael Yang and Nabil Omi and Negin Raoof and Shanda Li and Terry Yue Zhuo and Wuwei Lin and Yiwei Dai and Yuxin Wang and Wenhao Chai and Shang Zhou and Dariush Wahdany and Ziyu She and Jiaming Hu and Zhikang Dong and Yuxuan Zhu and Sasha Cui and Ahson Saiyed and Arinbjörn Kolbeinsson and Jesse Hu and Christopher Michael Rytting and Ryan Marten and Yixin Wang and Alex Dimakis and Andy Konwinski and Ludwig Schmidt},
      year={2026},
      eprint={2601.11868},
      archivePrefix={arXiv},
      primaryClass={cs.SE},
      url={https://arxiv.org/abs/2601.11868}, 
}

@misc{trivedi2024appworldcontrollableworldapps,
      title={AppWorld: A Controllable World of Apps and People for Benchmarking Interactive Coding Agents}, 
      author={Harsh Trivedi and Tushar Khot and Mareike Hartmann and Ruskin Manku and Vinty Dong and Edward Li and Shashank Gupta and Ashish Sabharwal and Niranjan Balasubramanian},
      year={2024},
      eprint={2407.18901},
      archivePrefix={arXiv},
      primaryClass={cs.SE},
      url={https://arxiv.org/abs/2407.18901}, 
}

@misc{liu2026skillforgeforgingdomainspecificselfevolving,
      title={SkillForge: Forging Domain-Specific, Self-Evolving Agent Skills in Cloud Technical Support}, 
      author={Xingyan Liu and Xiyue Luo and Linyu Li and Ganghong Huang and Jianfeng Liu and Honglin Qiao},
      year={2026},
      eprint={2604.08618},
      archivePrefix={arXiv},
      primaryClass={cs.IR},
      doi={https://doi.org/10.1145/3805712.3808466},
      url={https://arxiv.org/abs/2604.08618}, 
}

@misc{madaan2023selfrefineiterativerefinementselffeedback,
      title={Self-Refine: Iterative Refinement with Self-Feedback}, 
      author={Aman Madaan and Niket Tandon and Prakhar Gupta and Skyler Hallinan and Luyu Gao and Sarah Wiegreffe and Uri Alon and Nouha Dziri and Shrimai Prabhumoye and Yiming Yang and Shashank Gupta and Bodhisattwa Prasad Majumder and Katherine Hermann and Sean Welleck and Amir Yazdanbakhsh and Peter Clark},
      year={2023},
      eprint={2303.17651},
      archivePrefix={arXiv},
      primaryClass={cs.CL},
      url={https://arxiv.org/abs/2303.17651}, 
}

@misc{xu2026agentskillslargelanguage,
      title={Agent Skills for Large Language Models: Architecture, Acquisition, Security, and the Path Forward}, 
      author={Renjun Xu and Yang Yan},
      year={2026},
      eprint={2602.12430},
      archivePrefix={arXiv},
      primaryClass={cs.MA},
      url={https://arxiv.org/abs/2602.12430}, 
}

@misc{jiang2026sokagenticskills,
      title={SoK: Agentic Skills -- Beyond Tool Use in LLM Agents}, 
      author={Yanna Jiang and Delong Li and Haiyu Deng and Baihe Ma and Xu Wang and Qin Wang and Guangsheng Yu},
      year={2026},
      eprint={2602.20867},
      archivePrefix={arXiv},
      primaryClass={cs.CR},
      url={https://arxiv.org/abs/2602.20867}, 
}

@misc{wang2022scienceworldagentsmarter5th,
      title={ScienceWorld: Is your Agent Smarter than a 5th Grader?}, 
      author={Ruoyao Wang and Peter Jansen and Marc-Alexandre Côté and Prithviraj Ammanabrolu},
      year={2022},
      eprint={2203.07540},
      archivePrefix={arXiv},
      primaryClass={cs.CL},
      url={https://arxiv.org/abs/2203.07540}, 
}

@misc{wang2025inducingprogrammaticskillsagentic,
      title={Inducing Programmatic Skills for Agentic Tasks}, 
      author={Zora Zhiruo Wang and Apurva Gandhi and Graham Neubig and Daniel Fried},
      year={2025},
      eprint={2504.06821},
      archivePrefix={arXiv},
      primaryClass={cs.CL},
      url={https://arxiv.org/abs/2504.06821}, 
}

@misc{chen2023teachinglargelanguagemodels,
      title={Teaching Large Language Models to Self-Debug}, 
      author={Xinyun Chen and Maxwell Lin and Nathanael Schärli and Denny Zhou},
      year={2023},
      eprint={2304.05128},
      archivePrefix={arXiv},
      primaryClass={cs.CL},
      url={https://arxiv.org/abs/2304.05128}, 
}

@misc{guo2025stabletoolbenchstablelargescalebenchmarking,
      title={StableToolBench: Towards Stable Large-Scale Benchmarking on Tool Learning of Large Language Models}, 
      author={Zhicheng Guo and Sijie Cheng and Hao Wang and Shihao Liang and Yujia Qin and Peng Li and Zhiyuan Liu and Maosong Sun and Yang Liu},
      year={2025},
      eprint={2403.07714},
      archivePrefix={arXiv},
      primaryClass={cs.CL},
      url={https://arxiv.org/abs/2403.07714}, 
}

\clearpage
\appendix

%

\appendix

\FloatBarrier
\section{Experimental Setup and Environment Shifts}
\label{app:evaluation-details}

\subsection{Models and Execution Harnesses}
\label{app:actors}

We evaluate Claude Opus~4.7 (\texttt{claude-opus-4-7}) and
Claude Haiku~4.5 (\texttt{claude-haiku-4-5}). Within each
environment, the models use the same execution harness, tools,
task instructions, stopping criteria, and scoring across skill
conditions. On \tb, both run as Claude Code agents through
Harbor\footnote{\url{https://github.com/harbor-framework/harbor}}
\citep{merrill2026terminalbenchbenchmarkingagentshard}.
Opus~4.7 also creates the skills and performs model-based
verification, repair, and rebuilding. Maintenance changes the
skill state while keeping the task-performing agent fixed.

An \emph{agent turn} is one model decision step in the harness.
We measure execution cost in turns, except on \scienceworld,
where we count environment steps. IRS allows 15 turns per run,
AppWorld 50 turns, and ScienceWorld 90 steps. TB-2 uses each
task's time limit without a turn cap. A run that reaches its
limit is scored on the output produced, and all consumed turns
or steps count toward cost. Runs with no scorable output receive
zero task performance.

\subsection{Skill Construction}
\label{app:skill-construction}

We create procedural and cache-heavy versions of each skill
from the same experience in $W_0$. Procedural skills explain
how to inspect the environment and determine current values.
Cache-heavy skills store more of the choices resolved during
authoring. Both use Opus~4.7 and the refinement procedure of
\citet{liu2026skillswild}, with a format-only version of
Anthropic's skill-creator\footnote{\url{https://github.com/anthropics/skills}}
as the authoring guide. The author tests its draft in $W_0$
before producing the skill and is instructed to exclude task
answers and assumptions about future changes.

The authoring-style instruction in
Figure~\ref{fig:authoring-style} is the part of the prompt
changed between variants. Each resulting skill is created once
and used unchanged by both models before and after the
environment change. The styles differ in the choices they
retain and the procedures they provide for finding current
information, as discussed in
Section~\ref{sec:specificity-capability}.

\par\medskip\noindent
\begin{minipage}{\columnwidth}
\centering
\begin{promptbox}{Procedural}{procframe}
Write the skill as a reusable \emph{procedure}. Describe how to inspect the
environment and derive task-specific values at run time rather than
hardcoding them. Use placeholders for instance-variable parameters and note
where the agent should read each value.
\end{promptbox}
\vspace{6pt}
\begin{promptbox}{Cache-heavy}{cacheframe}
\emph{Cache} the resolved solution as directly as possible. Wherever you
resolved an environment-specific decision, write the resolved value into the
skill as a literal and include the working commands or routine verbatim. Do
not leave discovery, inspection, or derivation steps; do not use placeholders
or fallbacks. Never write the task answer or any value computed from this
instance's input.
\end{promptbox}
\vspace{6pt}
\captionof{figure}{\textbf{Authoring-style instructions (condensed).}}
\label{fig:authoring-style}
\end{minipage}\par\medskip

\subsection{Evaluation Tasks and Metrics}
\label{app:evaluation-populations}

Table~\ref{tab:evaluation-populations} summarizes the tasks used
to evaluate skill reuse before and after an environment change.
Within each environment, the same evaluation tasks are used
with and without the skill.

\par\medskip\noindent
\begin{minipage}{\columnwidth}
\centering
\captionof{table}{\textbf{Evaluation tasks and metrics.} IRS uses ten applicants for each of five form families.}
\label{tab:evaluation-populations}
\label{tab:harness-settings}
\scriptsize
\setlength{\tabcolsep}{2.5pt}
\renewcommand{\arraystretch}{1.05}
\begin{tabular}{@{}llll@{}}
\toprule
\textbf{Environment} & \textbf{Tasks} & \textbf{Score} & \textbf{Cost} \\
\midrule
\tb           & \shortstack[l]{38 task--change\\pairs} & pass rate & turns \\
\irs          & 50 (5 forms $\times$ 10) & field accuracy & turns \\
\appworld     & 51 tasks                & task success   & turns \\
\scienceworld & 41 variations           & score          & steps \\
\bottomrule
\end{tabular}
\end{minipage}
\par\medskip

TB-2 includes 38 environment changes across 30 tasks. Each
change is applied separately to the original task environment.
We compare each changed version with its corresponding $W_0$
runs and average over the 38 task--change pairs. Most tasks have
one run per condition; repeated runs, where available, are
averaged, with up to nine runs for some tasks.

On IRS, field accuracy is the fraction of in-scope fields
matching the reference form. Results average the five
form-family means, each based on ten applicants. Applicants
A--C support skill creation, repair tests, and candidate
screening; D--E provide further acceptance checks; F--O are
used to report task performance and execution cost.

AppWorld uses 51 tasks from the \texttt{test\_normal} split
\citep{trivedi2024appworldcontrollableworldapps}; a task succeeds
when all its evaluation tests pass. ScienceWorld uses 41
variations of \texttt{test-conductivity}
\citep{wang2022scienceworldagentsmarter5th}, excluding variations
used for skill creation. Each starts at least two hops from the
workshop. We divide simulator scores by 100 to obtain values
in $[0,1]$.

Each IRS applicant, AppWorld task, and ScienceWorld variation
is run once under each experimental condition. We compute
performance gains and relative cost reductions using
Section~\ref{sec:formulation}, with the no-skill agent in the
same environment as the baseline.

\subsection{Environment Transitions}
\label{app:shift-construction}

\begin{figure*}[!t]
\centering
\begin{minipage}[t]{0.495\textwidth}
\begin{driftcard}{\tb\ \textperiodcentered\ \texttt{git-multibranch}}{Git push-to-deploy with Nginx}
\envpair{\texttt{nginx.conf} loads \texttt{sites-enabled/}, which links the default site}
        {The \texttt{sites-enabled} include and link are removed}
\persisted{Cache-heavy skill retains}
  {``Nginx default site config path: \texttt{/etc/nginx/sites-available/default}
   (already symlinked from \texttt{sites-enabled/default})''\par
   {\itshape\color{inkgray}The site is written to a file Nginx no longer loads.}}
\recoverable{\texttt{nginx.conf} lists what Nginx loads; the procedural skill
says to ``write the block to the file included by \texttt{nginx.conf}''.}
\end{driftcard}
\end{minipage}\hfill
\begin{minipage}[t]{0.495\textwidth}
\begin{driftcard}{\irs\ \textperiodcentered\ Form W-9}{2018 $\rightarrow$ 2024 revision}
\envpair{Line 1 (name) is field \texttt{f1\_1[0]}}
        {Line 1 (name) is field \texttt{f1\_01[0]}; lines 1--9 renumbered}
\persisted{Cache-heavy skill retains}
  {\texttt{\_set\_text(writer, "\ldots f1\_1[0]", line1\_name)}\par
   {\itshape\color{inkgray}The field no longer exists, so the write is silently
   ignored.}}
\recoverable{Field names and printed labels on the current form; the
procedural skill says to ``enumerate \texttt{reader.get\_fields()} to confirm
the leaf names''.}
\end{driftcard}
\end{minipage}

\vspace{6pt}

\begin{minipage}[t]{0.495\textwidth}
\begin{driftcard}{\appworld\ \textperiodcentered\ login capability}{1 of 3 renamed endpoints}
\envpair{Credentials come from \texttt{show\_\allowbreak account\_\allowbreak passwords}}
        {The same endpoint is now \texttt{show\_\allowbreak credential\_\allowbreak vault}}
\persisted{Both skill variants retain}
  {\texttt{apis.supervisor.show\_account\_passwords()}\par
   {\itshape\color{inkgray}The login routine reads credentials from the old
   endpoint.}}
\recoverable{The renamed endpoint is listed in the current API
documentation.}
\end{driftcard}
\end{minipage}\hfill
\begin{minipage}[t]{0.495\textwidth}
\begin{driftcard}{\scienceworld\ \textperiodcentered\ \texttt{test-conductivity}}{relocation}
\envpair{The aluminum foil is in the room the task names}
        {The foil is in the agent's starting room; the task text is unchanged}
\persisted{Cache-heavy skill retains}
  {``Phase 1 --- Navigate to TARGET\_ROOM''\par
   {\itshape\color{inkgray}TARGET\_ROOM is the room named after ``located
   around the''.}}
\recoverable{The foil is in plain view in the starting room.}
\end{driftcard}
\end{minipage}
\caption{
\textbf{Examples of environment transitions.}
Each card shows one transition, the content retained from $W_0$ that
the change invalidates, and where the changed information can still be found in
$W_1$.
Quoted text is taken from the skills used in our experiments.
}
\label{fig:transition-examples}
\end{figure*}

The task objective, scoring criterion, and skill state remain fixed
while the environment changes. Figure~\ref{fig:transition-examples}
shows the affected skill content and the information available
in the current environment.

On TB-2, each change affects a package, library behavior,
configuration, data schema, or task file in the task container.
On IRS, $W_1$ is a later revision\footnote{\scriptsize\url{https://www.irs.gov/prior-year-forms-and-instructions}}
of the same form: W-9 2018$\rightarrow$2024, W-4
2019$\rightarrow$2020, 1099-MISC 2019$\rightarrow$2024,
and 8606 and 8949 2017$\rightarrow$2025. The revisions rename,
renumber, move, add, or remove fields. Reference forms use the
field layout of the revision being evaluated.

On AppWorld, three endpoints are renamed together:
\texttt{show\_account\_passwords},
\texttt{show\_transactions}, and \texttt{show\_profile}.
The callable API, route names, and documentation are updated
consistently. On ScienceWorld, ordinary simulator actions move
the target object from the room named in the task to the
agent's starting room. A matched sham setup makes the same
trip and returns the object before execution.

\FloatBarrier
\section{Additional Staleness Results}
\label{app:extended-staleness}

\subsection{Procedural Skills}
\label{app:procedural-broad-value}
\label{app:full-factorial}

\begin{table*}[!t]
\centering
\caption{
\textbf{Performance and cost with procedural skills before and after the change.}
Cells follow Table~\ref{tab:broad-value}; colors describe the direction of the reported difference, not statistical significance.
}
\label{tab:procedural-broad-value}
\label{tab:additional-factorial}

\begingroup
\scriptsize
\setlength{\tabcolsep}{2.15pt}
\renewcommand{\arraystretch}{1.04}

\begin{tabular}{@{}llccc ccc@{}}
\toprule
&
&
\multicolumn{3}{c}{\textbf{Task performance}}
&
\multicolumn{3}{c}{\textbf{Execution cost}}
\\
\cmidrule(lr){3-5}
\cmidrule(lr){6-8}

\textbf{Environment}
&
\textbf{Model}
&
\shortstack{\textbf{Healthy $W_0$}\\[-1pt]No $\rightarrow$ Skill}
&
\shortstack{\textbf{Shifted $W_1$}\\[-1pt]No $\rightarrow$ Skill}
&
\shortstack{\textbf{Loss}\\[-1pt]$L_U$}
&
\shortstack{\textbf{Healthy $W_0$}\\[-1pt]No $\rightarrow$ Skill}
&
\shortstack{\textbf{Shifted $W_1$}\\[-1pt]No $\rightarrow$ Skill}
&
\shortstack{\textbf{Loss}\\[-1pt]$L_S$}
\\
\midrule

\multirow{2}{*}{\tb}
& Opus
& \skillpair{.921}{.932}{+1.1\,\mathrm{pp}}
& \skillpair{.895}{.974}{+7.9\,\mathrm{pp}}
& \improvecell{-6.8\,\mathrm{pp}}
& \skillpair{40.7}{26.5}{34.9\%}
& \skillpair{37.6}{38.9}{-3.4\%}
& \losscell{38.3\,\mathrm{pp}}
\\

& Haiku
& \skillpair{.447}{.848}{+40.1\,\mathrm{pp}}
& \skillpair{.447}{.711}{+26.3\,\mathrm{pp}}
& \losscell{13.8\,\mathrm{pp}}
& \skillpair{66.1}{46.5}{29.7\%}
& \skillpair{68.9}{60.5}{12.1\%}
& \losscell{17.6\,\mathrm{pp}}
\\

\addlinespace[1.5pt]

\multirow{2}{*}{\irs}
& Opus
& \skillpair{.772}{.960}{+18.9\,\mathrm{pp}}
& \skillpair{.856}{.919}{+6.2\,\mathrm{pp}}
& \losscell{12.6\,\mathrm{pp}}
& \skillpair{10.0}{3.5}{65.0\%}
& \skillpair{10.3}{10.2}{1.0\%}
& \losscell{64.0\,\mathrm{pp}}
\\

& Haiku
& \skillpair{.007}{.931}{+92.4\,\mathrm{pp}}
& \skillpair{.046}{.482}{+43.6\,\mathrm{pp}}
& \losscell{48.8\,\mathrm{pp}}
& \skillpair{13.1}{4.6}{65.0\%}
& \skillpair{12.3}{7.2}{41.6\%}
& \losscell{23.4\,\mathrm{pp}}
\\

\addlinespace[1.5pt]

\multirow{2}{*}{\appworld}
& Opus
& \skillpair{.863}{.961}{+9.8\,\mathrm{pp}}
& \skillpair{.863}{.922}{+5.9\,\mathrm{pp}}
& \losscell{3.9\,\mathrm{pp}}
& \skillpair{20.2}{13.4}{33.8\%}
& \skillpair{22.6}{17.5}{22.7\%}
& \losscell{11.1\,\mathrm{pp}}
\\

& Haiku
& \skillpair{.490}{.569}{+7.8\,\mathrm{pp}}
& \skillpair{.471}{.627}{+15.7\,\mathrm{pp}}
& \improvecell{-7.8\,\mathrm{pp}}
& \skillpair{33.1}{22.3}{32.7\%}
& \skillpair{33.4}{25.7}{22.9\%}
& \losscell{9.8\,\mathrm{pp}}
\\

\addlinespace[1.5pt]

\multirow{2}{*}{\scienceworld}
& Opus
& \skillpair{1.000}{1.000}{+0.0\,\mathrm{pp}}
& \skillpairneg{1.000}{.859}{-14.1\,\mathrm{pp}}
& \losscell{14.1\,\mathrm{pp}}
& \skillpair{17.3}{15.3}{11.6\%}
& \skillpair{19.3}{28.5}{-47.7\%}
& \losscell{59.2\,\mathrm{pp}}
\\

& Haiku
& \skillpair{.859}{1.000}{+14.1\,\mathrm{pp}}
& \skillpair{.927}{1.000}{+7.3\,\mathrm{pp}}
& \losscell{6.8\,\mathrm{pp}}
& \skillpair{52.3}{17.8}{66.1\%}
& \skillpair{55.0}{25.6}{53.4\%}
& \losscell{12.6\,\mathrm{pp}}
\\

\bottomrule
\end{tabular}
\endgroup
\end{table*}

Table~\ref{tab:procedural-broad-value} complements the cache-heavy
results in Table~\ref{tab:broad-value}. With procedural skills,
relative cost reduction falls in all eight model--environment
comparisons, and the performance gain over no skill falls in
six. Together, the tables support the comparisons of authoring
styles and model responses in Section~\ref{sec:runtime-results}.

\subsection{Comparing Authoring Styles}
\label{app:specificity-statistics}

Table~\ref{tab:specificity-statistics} compares the two styles'
original cost reduction and subsequent performance loss.
Positive values mean that cache-heavy skills save more cost
in $W_0$ or lose more performance gain after the change.
The IRS entries report the applicant-level paired analysis.

Confidence intervals use a paired bootstrap, resampling
38 TB-2 task--change pairs 10{,}000 times and 50 IRS
form--applicant pairs 20{,}000 times. Each resampled pair keeps
its before/after and skill/no-skill results together. TB-2 cost
ratios are recomputed within each resample. The table gives
95\% percentile intervals; the colors in
Table~\ref{tab:factorial-value} show the direction of the
reported differences.

\par\medskip\noindent
\begin{minipage}{\columnwidth}
\centering
\captionof{table}{\textbf{Paired authoring-style comparisons.}
Entries are cache-heavy minus procedural, in percentage points,
with 95\% paired-bootstrap intervals.}
\label{tab:specificity-statistics}
\scriptsize
\setlength{\tabcolsep}{2.4pt}
\renewcommand{\arraystretch}{1.05}

\begin{tabular}{@{}llcc@{}}
\toprule
\textbf{Env.}
&
\textbf{Model}
&
\shortstack{\textbf{Cost reduction in $W_0$}\\Cache $-$ Proc.}
&
\shortstack{\textbf{Performance loss}\\Cache $-$ Proc.}
\\
\midrule
\tb  & Opus  & $+31.5$ $[+22.0,+46.1]$ & $+22.9$ $[+11.0,+36.1]$ \\
\tb  & Haiku & $+31.2$ $[+18.2,+45.3]$ & $+17.5$ $[-7.9,+42.1]$ \\
\irs & Opus  & $+19.0$ $[+13.5,+26.7]$ & $+18.7$ $[+10.2,+27.8]$ \\
\irs & Haiku & $-0.9$ $[-9.6,+10.2]$   & $+16.8$ $[+6.8,+27.2]$ \\
\bottomrule
\end{tabular}
\end{minipage}
\par\medskip

\FloatBarrier
\section{How Agents Respond to Stale Skill States}
\label{app:runtime-annotations}

We analyze whether an agent inspects the current form and
whether it subsequently writes to an obsolete field identifier.
The analysis covers 200 IRS trajectories with stale skill states:
50 for each model and authoring style, spanning five forms and
ten applicants per form. We compute both indicators from
harness records after execution, without model-based annotation.

Table~\ref{tab:runtime-response} gives the outcomes for
inspecting and non-inspecting trajectories. Across both
styles, Opus inspects in 73 of 100 trajectories and Haiku
in 80. Obsolete writes occur after inspection in 7 of those
73 Opus trajectories and 36 of the 80 Haiku trajectories.
The difference helps explain why inspecting the current form
does not always lead to a successful correction.

\par\medskip\noindent
\begin{minipage}{\columnwidth}
\centering
\captionof{table}{
\textbf{IRS runtime response under stale-skill reuse.}
Each row covers 50 trajectories (5 forms $\times$ 10 applicants).
\emph{Inspects}: reads the current form before its first final program.
\emph{Obsolete write}: writes to a field name that exists only in $W_0$.
Accuracy and turns are pooled over forms; within-form contrasts are in
Table~\ref{tab:runtime-main}.
}
\label{tab:runtime-response}
\scriptsize
\setlength{\tabcolsep}{2.3pt}
\renewcommand{\arraystretch}{1.05}
\begin{tabular}{@{}llcccc@{}}
\toprule
& & & \shortstack{\textbf{Obsolete}\\\textbf{write}} & \textbf{Accuracy} & \textbf{Turns} \\
\textbf{Model} & \textbf{Skill} & \textbf{Inspects} &
insp.\ / not & insp.\ / not & insp.\ / not \\
\midrule
Opus  & Proc.  & 43/50 & 5/43 \,/\, 6/7   & .966 / .630 & 10.9 / 6.3 \\
Opus  & Cache  & 30/50 & 2/30 \,/\, 11/20 & .918 / .452 & 12.2 / 1.2 \\
\addlinespace[1.5pt]
Haiku & Proc.  & 41/50 & 14/41 \,/\, 5/9  & .516 / .326 & 8.3 / 2.1 \\
Haiku & Cache  & 39/50 & 22/39 \,/\, 6/11 & .312 / .280 & 8.4 / 1.9 \\
\bottomrule
\end{tabular}
\end{minipage}
\par\medskip

\paragraph{Behavioral indicators.}
A trajectory counts as inspecting when an exploratory step
before its first final program reads the current form's
fields or annotations through pypdf, or its printed text
through pdfplumber. Reading the applicant record alone does
not count. An obsolete write targets a field present in
$W_0$ and absent from $W_1$. Such writes also occur in 3 of
50 Opus no-skill trajectories and 7 of 50 Haiku no-skill
trajectories. The indicator records the use of an obsolete
identifier, without attributing every occurrence to the skill.

\paragraph{Within-form comparisons.}
Inspection rates differ by form
(Table~\ref{tab:runtime-by-form}). Table~\ref{tab:runtime-main}
therefore compares inspecting and non-inspecting trajectories
within each form, then averages over forms containing both
groups. A form receives weight $n_1n_0/(n_1+n_0)$, where
$n_1$ and $n_0$ are the two group sizes. Forms with more
observations in both groups receive more weight.
These comparisons describe the association between inspection
and outcomes for the same form.

\par\medskip\noindent
\begin{minipage}{\columnwidth}
\centering
\captionof{table}{
\textbf{Inspecting trajectories by form}, out of 10 per form.
}
\label{tab:runtime-by-form}
\scriptsize
\setlength{\tabcolsep}{3pt}
\renewcommand{\arraystretch}{1.05}
\begin{tabular}{@{}llccccc@{}}
\toprule
\textbf{Model} & \textbf{Skill} & W-9 & W-4 & 1099 & 8606 & 8949 \\
\midrule
Opus  & Proc.  & 10 & 10 & 10 & 9 & 4 \\
Opus  & Cache  & 10 & 3  & 8  & 9 & 0 \\
\addlinespace[1.5pt]
Haiku & Proc.  & 10 & 7  & 9  & 7 & 8 \\
Haiku & Cache  & 8  & 9  & 8  & 7 & 7 \\
\bottomrule
\end{tabular}
\end{minipage}
\par\medskip

\paragraph{Why lower cost can accompany worse performance.}
For Opus with cache-heavy skills, accuracy falls 12.5 percentage
points below no skill while turns fall by 24.6\%.
The 20 non-inspecting trajectories contribute $-16.7$ points
to the accuracy difference and $+37.1$ points to cost reduction;
the 30 inspecting trajectories contribute $+4.2$ and $-12.5$
points, respectively. These weighted contributions sum to the
overall differences. The inexpensive, non-inspecting runs
account for the aggregate saving, while the more successful
inspecting runs use 2.2 more turns on average than the no-skill
runs for their forms.

\FloatBarrier
\section{Maintenance Implementation}
\label{app:maintenance}
\label{app:lifecycle}

\subsection{Constructing and Replaying Checks}
\label{app:profile-construction}
\label{app:profile-adapters}
\label{app:detection}

The execution profile records the environment properties a
skill uses, so the maintainer can later check whether its
retained choices still apply. Check construction extracts
references from the skill and operations from successful
executions. Shared observation routines turn these references
and operations into executable checks, selecting what to test
automatically for each skill.

On TB-2, rule-based extraction identifies paths, commands,
and identifiers. Unresolved identifiers are discarded, as are
checks that vary across three $W_0$ replays. On IRS, the
skill's fill routine runs on synthetic marker inputs. The
profile records whether each write reaches a field, its
location and printed label, and any checkbox on-state;
applicant values are excluded. On AppWorld and ScienceWorld,
checks come from the API calls and actions issued by code
from successful recorded executions. Construction and replay
make no model calls.

Each record contains an executable check $q_j$, its output
$e_j^{(0)}$ in $W_0$, and a source reference $p_j$
(Equation~\ref{eq:execution-profile}). Outputs are Booleans,
strings, or structured observations. Source references
identify the instruction, field name, or recorded operation
from which the check was built. For a saved script path,
the check tests whether the file exists, the recorded result
states that it existed in $W_0$, and the source reference
locates the instruction that uses it. A later failure gives
verification both evidence of change and content to examine.

This construction supports the observation routines provided
for each kind of environment. A new kind of resource needs
a routine that can observe its relevant properties; skill
references then determine where to apply it. Before reuse,
the maintainer can run the checks and compare their results
with the recorded observations. Each maintenance evaluation
starts from the original skill and applies one environment
change, followed by one replay. The controls are described
in Appendix~\ref{app:monitoring-analysis}.

\subsection{Verification and Updates}
\label{app:verification}
\label{app:update-details}
\label{app:information-boundary}

\begin{table*}[!t]
\centering
\caption{\textbf{Inputs to maintenance.}
The maintainer receives observed changes and execution feedback;
no external changelog is supplied.}
\label{tab:information-boundary}
\scriptsize
\setlength{\tabcolsep}{4pt}
\renewcommand{\arraystretch}{1.12}
\begin{tabular}{@{}p{.13\textwidth}p{.80\textwidth}@{}}
\toprule
\textbf{Stage} & \textbf{Inputs and feedback} \\
\midrule
Construction & Skill references, successful executions, and observations in $W_0$; IRS field checks use synthetic marker inputs. \\
Replay & Recorded checks and reference observations, evaluated against the current environment without model calls. \\
Verification & Skill content, original/current observations, and execution statistics from the recorded runs, as described below. \\
Repair & Affected content, detected differences or verifier report, and environment access or execution feedback. TB-2 retries receive failing test output, which can include expected values. IRS supplies the blank current form, not completed evaluation forms. \\
Rebuilding & Current-environment authoring inputs described in the text; no stale skill. \\
Acceptance & Candidate task performance and execution cost on validation runs, together with the recorded reference levels. \\
\bottomrule
\end{tabular}
\end{table*}

\paragraph{Verification.}
A changed observation is checked against the skill content
that depends on it. An invalid call or changed action can
identify the responsible component directly, as in AppWorld
and ScienceWorld. On TB-2 and IRS, Opus~4.7 judges the skill,
changed checks, and original observations. Three judgments
are sampled separately from the same prompt and evidence;
at least two must support a conflict. The most confident
agreeing judgment supplies the location and explanation
passed to repair.

The verifier also receives the skill's mean turns before and
after the change from recorded runs. On TB-2 it additionally
receives token counts and the change in no-skill turns; IRS
includes a trajectory-change signal. Table~\ref{tab:information-boundary}
summarizes these inputs. The no-skill comparison supplies
additional experimental evidence for TB-2 verification.

\paragraph{Repair.}
The maintainer receives the affected content and conflict
report. It is instructed to correct outdated references,
retain useful content and the authoring style, and keep the
skill reusable across task inputs. Candidate tests provide
feedback for further edits. On TB-2, retries receive
validation accuracy, turns, and up to twelve failing lines
from the task tests. IRS supplies the blank current form's
field names and printed labels. AppWorld tests return
program output from fresh environment instances; ScienceWorld
tests execute the candidate with Haiku and return scores,
steps, and the last actions of the weakest episode.

\paragraph{Rebuilding.}
Rebuilding reruns the original authoring procedure in $W_1$
without the stale skill state. TB-2 uses exploration of the task
container without access to its tests. IRS uses successful
no-skill runs on training applicants; AppWorld uses interaction
traces from no-skill runs on training templates.
ScienceWorld uses reference action sequences from the
simulator's training variations and tests candidates in
the current environment. The authoring style remains fixed.

\paragraph{Acceptance.}
\label{app:acceptance-rules}
We validate candidates using task performance and execution cost.
The performance threshold is the lower of the healthy and stale
skill's validation scores, and the cost target is the healthy
skill's validation cost.
A candidate meeting both targets is accepted.

When these targets cannot be met, a fallback accepts a candidate
whose performance is no more than one standard error below the
stale score, whose cost does not exceed stale execution cost,
and which improves at least one of these measures.
The same criteria apply to repair and rebuilding.
Selective accepts a repair first, attempts rebuilding if repair
is unsuccessful, and retains the existing state if neither
candidate is accepted.
All acceptance statistics are computed from development runs,
separately from reported evaluation.

\subsection{Verification and Repair Prompts}
\label{app:prompts}

\begin{figure*}[!t]
\centering
\begin{minipage}[t]{.49\textwidth}
\vspace{0pt}
\begin{promptbox}{Verification prompt (TB-2)}{procframe}
\small
YOUR JOB is to AUDIT the alarm, not rubber-stamp it. You may
rethink, reinterpret, accept or REJECT.

\medskip
\textit{[\ldots]}
A shift that costs only efficiency is still real drift: do NOT return
task\_noise merely because the task would still eventually pass.

\medskip
\textit{[\ldots]}
Only return real\_drift when the evidence shows a structural or
semantic change in the environment.
\end{promptbox}
\end{minipage}\hfill
\begin{minipage}[t]{.49\textwidth}
\vspace{0pt}
\begin{promptbox}{Repair prompt (TB-2)}{cacheframe}
\small
EFFICIENCY -- it must again SUPPLY what it used to supply. Do not
simply delete the stale specifics and tell the agent to work it
out: that restores correctness while throwing away the reason
the skill existed.

\medskip
\textit{[\ldots]}
Keep the skill's structure and voice; edit only what is stale.

Never write a value that a grader checks.

Output the COMPLETE updated file and nothing else. No fences,
no commentary.
\end{promptbox}
\end{minipage}
\caption{\textbf{Selected instructions from the verification and repair prompts.}
Text is quoted from the TB-2 system prompts; brackets mark
omitted passages. Full prompts accompany the code.}
\label{fig:maintenance-prompts}
\end{figure*}

Verification determines whether an observed change conflicts
with content the skill relies on. A confirmed conflict is
linked to the affected instruction, code, or recorded operation,
giving repair both the evidence and a location to examine.
The repair model uses this information to update outdated
content while preserving useful procedures and keeping the
skill reusable across task inputs. The updated skill is tested
before acceptance. Figure~\ref{fig:maintenance-prompts} shows
selected instructions for these two steps; full prompts
accompany the code.

\subsection{Trigger Evaluation}
\label{app:monitoring-analysis}
\label{app:profile-ablation}

The trigger is evaluated on changes that invalidate skill
content and controls that leave it valid. TB-2 reference
labels are derived from the retained content and the
controlled change. A conflict invalidates a path, command,
or literal used by the skill. Table~\ref{tab:trigger-coverage}
separates these conflicts from equivalent-change controls.

\par\medskip\noindent
\begin{minipage}{\columnwidth}
\centering
\captionof{table}{\textbf{TB-2 trigger outcomes.}
Each style is evaluated on 38 transitions; the conflict rows
count those labeled as invalidating retained skill content.
The separate control suite contains 62 equivalent changes.
Verification counts use the Opus maintenance runs.}
\label{tab:trigger-coverage}
\scriptsize
\setlength{\tabcolsep}{3pt}
\renewcommand{\arraystretch}{1.08}
\begin{tabular}{@{}lrrr@{}}
\toprule
\textbf{Cases} & \textbf{Total} & \textbf{Flagged} & \textbf{Verified} \\
\midrule
Procedural conflicts & 22 & 15 & 9 \\
Cache-heavy conflicts & 29 & 19 & 19 \\
\addlinespace[2pt]
Equivalent-change controls & 62 & 7 & 2 \\
\bottomrule
\end{tabular}
\end{minipage}
\par\medskip

Across both styles, replay flags 34 of 51 labeled conflicts
and verification supports 28. The 17 replay misses involve
byte-level changes outside the recorded properties. Every
verified conflict is linked to an affected artifact.

Among the 62 equivalent-change controls, replay flags seven
and verification rejects five; the other two lead to
unnecessary procedural-skill repairs. On IRS, verification
accepts all ten conflicting transitions and rejects four
other form revisions with unchanged field names. Replaying
the unchanged AppWorld environment flags none of its 51
tasks. These controls show why a changed observation needs
verification before an update is attempted.

\FloatBarrier
\section{Maintenance Results and Cost}
\label{app:maintenance-protocol}
\label{app:maintenance-references}

\subsection{Policy Evaluation}
\label{app:opus-maintenance}
\label{app:tb-maintenance}
\label{app:irs-maintenance}
\label{app:appworld-maintenance}
\label{app:scienceworld-maintenance}
\label{app:maintenance-statistics}
\label{app:decision-paths}

\begin{table*}[!t]
\centering
\caption{\textbf{Maintenance results by policy.}
Parentheses show the performance gain (percentage points) or cost
reduction relative to no skill in $W_1$. Green marks improvement
over No maintenance; red marks negative performance gain under
No maintenance. Colors describe the reported means.
Cost is in agent turns, except for ScienceWorld steps.
$^\dagger$Archived runs matched by run name.
$^\ddagger$The candidate failed validation, so the policy keeps
the stale skill.
$^\S$Rebuilding covers 30 of 41 variations; these entries show
absolute outcomes on that subset.}
\label{tab:maint-policies}
\label{tab:opus-maintenance}
\label{tab:tb-full-maintenance}
\label{tab:irs-full-maintenance}

\begingroup
\scriptsize
\setlength{\tabcolsep}{3.0pt}
\renewcommand{\arraystretch}{1.04}

\begin{tabular}{@{}lllccccc@{}}
\toprule
\textbf{Environment} & \textbf{Model} & \textbf{Skill} & \textbf{No skill} &
\shortstack{\textbf{No}\\\textbf{maint.}} & \shortstack{\textbf{Repair}\\\textbf{only}} & \shortstack{\textbf{Rebuild}\\\textbf{only}} & \textbf{Selective} \\
\midrule
\multicolumn{8}{@{}l}{\textit{Task outcome}} \\
\multirow{4}{*}{\tb} & Opus & Cache & .895
  & \maintnegcell{.781}{-11.4\,\mathrm{pp}} & \maintgaincell{.820}{-7.5\,\mathrm{pp}} & \maintgaincell{.895}{+0.0\,\mathrm{pp}} & \maintgaincell{\textbf{.886}}{-0.9\,\mathrm{pp}} \\
 & Opus & Proc. & .895
  & \skillval{.974}{+7.9\,\mathrm{pp}} & \skillval{.965}{+7.0\,\mathrm{pp}} & \skillval{.974}{+7.9\,\mathrm{pp}} & \skillval{\textbf{.965}}{+7.0\,\mathrm{pp}} \\
 & Haiku & Cache & .447
  & \skillval{.561}{+11.4\,\mathrm{pp}} & \maintgaincell{.649}{+20.2\,\mathrm{pp}} & \maintgaincell{.610}{+16.2\,\mathrm{pp}}$^\dagger$ & \maintgaincell{\textbf{.649}}{+20.2\,\mathrm{pp}} \\
 & Haiku & Proc. & .447
  & \skillval{.711}{+26.3\,\mathrm{pp}} & \maintgaincell{.798}{+35.1\,\mathrm{pp}} & \maintgaincell{.746}{+29.8\,\mathrm{pp}}$^\dagger$ & \maintgaincell{\textbf{.798}}{+35.1\,\mathrm{pp}} \\
\addlinespace[1.5pt]
\multirow{4}{*}{\irs} & Opus & Cache & .856
  & \maintnegcell{.731}{-12.5\,\mathrm{pp}} & \maintgaincell{.981}{+12.4\,\mathrm{pp}} & \maintgaincell{.987}{+13.0\,\mathrm{pp}} & \maintgaincell{\textbf{.981}}{+12.4\,\mathrm{pp}} \\
 & Opus & Proc. & .856
  & \skillval{.919}{+6.2\,\mathrm{pp}} & \maintgaincell{.959}{+10.2\,\mathrm{pp}} & \maintgaincell{.949}{+9.2\,\mathrm{pp}} & \maintgaincell{\textbf{.951}}{+9.5\,\mathrm{pp}} \\
 & Haiku & Cache & .046
  & \skillval{.305}{+25.9\,\mathrm{pp}} & \maintgaincell{.975}{+92.9\,\mathrm{pp}} & \maintgaincell{.942}{+89.6\,\mathrm{pp}} & \maintgaincell{\textbf{.975}}{+92.9\,\mathrm{pp}} \\
 & Haiku & Proc. & .046
  & \skillval{.482}{+43.6\,\mathrm{pp}} & \maintgaincell{.802}{+75.6\,\mathrm{pp}} & \maintgaincell{.816}{+77.0\,\mathrm{pp}} & \maintgaincell{\textbf{.799}}{+75.3\,\mathrm{pp}} \\
\addlinespace[1.5pt]
\multirow{2}{*}{\appworld} & Haiku & Cache & .471
  & \skillval{.627}{+15.7\,\mathrm{pp}} & \maintgaincell{.667}{+19.6\,\mathrm{pp}} & \skillval{.627}{+15.7\,\mathrm{pp}}$^\ddagger$ & \maintgaincell{\textbf{.667}}{+19.6\,\mathrm{pp}} \\
 & Haiku & Proc. & .471
  & \skillval{.627}{+15.7\,\mathrm{pp}} & \skillval{.569}{+9.8\,\mathrm{pp}} & \skillval{.569}{+9.8\,\mathrm{pp}} & \skillval{\textbf{.569}}{+9.8\,\mathrm{pp}} \\
\addlinespace[1.5pt]
\multirow{2}{*}{\scienceworld} & Haiku & Cache & .927
  & \skillval{1.000}{+7.3\,\mathrm{pp}} & \skillval{1.000}{+7.3\,\mathrm{pp}} & .970$^\S$ & \skillval{\textbf{1.000}}{+7.3\,\mathrm{pp}} \\
 & Haiku & Proc. & .927
  & \skillval{1.000}{+7.3\,\mathrm{pp}} & \skillval{1.000}{+7.3\,\mathrm{pp}}$^\ddagger$ & \skillval{1.000}{+7.3\,\mathrm{pp}} & \skillval{\textbf{1.000}}{+7.3\,\mathrm{pp}} \\
\midrule
\multicolumn{8}{@{}l}{\textit{Execution cost}} \\
\multirow{4}{*}{\tb} & Opus & Cache & 37.6
  & \skillval{24.2}{35.6\%} & \maintgaincell{20.8}{44.7\%} & \maintgaincell{21.8}{42.2\%} & \maintgaincell{\textbf{19.5}}{48.0\%} \\
 & Opus & Proc. & 37.6
  & \skillval{38.9}{-3.4\%} & \maintgaincell{29.2}{22.3\%} & \maintgaincell{30.7}{18.5\%} & \maintgaincell{\textbf{28.9}}{23.1\%} \\
 & Haiku & Cache & 68.9
  & \skillval{38.9}{43.5\%} & \maintgaincell{32.1}{53.4\%} & \maintgaincell{35.3}{48.7\%}$^\dagger$ & \maintgaincell{\textbf{32.1}}{53.4\%} \\
 & Haiku & Proc. & 68.9
  & \skillval{60.5}{12.1\%} & \maintgaincell{59.7}{13.3\%} & \maintgaincell{56.7}{17.7\%}$^\dagger$ & \maintgaincell{\textbf{59.7}}{13.3\%} \\
\addlinespace[1.5pt]
\multirow{4}{*}{\irs} & Opus & Cache & 10.3
  & \skillval{7.8}{24.6\%} & \maintgaincell{3.8}{63.2\%} & \maintgaincell{1.8}{82.8\%} & \maintgaincell{\textbf{3.8}}{63.2\%} \\
 & Opus & Proc. & 10.3
  & \skillval{10.2}{1.0\%} & \maintgaincell{7.4}{28.2\%} & \maintgaincell{4.2}{59.2\%} & \maintgaincell{\textbf{5.5}}{46.8\%} \\
 & Haiku & Cache & 12.3
  & \skillval{7.0}{43.1\%} & \maintgaincell{4.7}{61.5\%} & \maintgaincell{4.3}{64.9\%} & \maintgaincell{\textbf{4.7}}{61.5\%} \\
 & Haiku & Proc. & 12.3
  & \skillval{7.2}{41.6\%} & \maintgaincell{6.4}{47.7\%} & \maintgaincell{6.5}{46.8\%} & \maintgaincell{\textbf{6.5}}{46.7\%} \\
\addlinespace[1.5pt]
\multirow{2}{*}{\appworld} & Haiku & Cache & 33.4
  & \skillval{28.7}{13.9\%} & \maintgaincell{23.3}{30.3\%} & \skillval{28.7}{13.9\%}$^\ddagger$ & \maintgaincell{\textbf{23.3}}{30.3\%} \\
 & Haiku & Proc. & 33.4
  & \skillval{25.7}{22.9\%} & \maintgaincell{24.8}{25.8\%} & \maintgaincell{24.2}{27.4\%} & \maintgaincell{\textbf{24.8}}{25.8\%} \\
\addlinespace[1.5pt]
\multirow{2}{*}{\scienceworld} & Haiku & Cache & 55.0
  & \skillval{28.0}{49.0\%} & \maintgaincell{18.6}{66.1\%} & 26.3$^\S$ & \maintgaincell{\textbf{18.6}}{66.1\%} \\
 & Haiku & Proc. & 55.0
  & \skillval{25.6}{53.4\%} & \skillval{25.6}{53.4\%}$^\ddagger$ & \maintgaincell{16.1}{70.7\%} & \maintgaincell{\textbf{16.1}}{70.7\%} \\
\bottomrule
\end{tabular}

\endgroup
\end{table*}

Table~\ref{tab:maint-policies} compares four policies.
No maintenance reuses the stale skill state unchanged.
Repair only and Rebuild only use the same trigger and accept
or reject their respective candidates through validation.
Selective attempts repair first, proceeds to rebuilding when
repair is unsuccessful, and otherwise retains the stale skill
if neither update is accepted.

We evaluate Selective by running this sequence and evaluating
the skill state returned by the controller.
For Haiku's procedural IRS skills, two controller runs yield
the same mean accuracy, .799.
Opus results cover TB-2 and IRS.
The cache-heavy ScienceWorld rebuilding result covers 30
variations, so comparisons requiring the same evaluation
tasks exclude that entry.

\paragraph{Performance and cost after maintenance.}
Table~\ref{tab:maint-paired} reports paired differences between
Selective and No maintenance. Cache-heavy TB-2 skills improve
on both measures. Procedural skills mainly improve cost under
Opus and performance under Haiku. On ScienceWorld, maintenance
reduces steps while preserving the observed perfect score.
AppWorld procedural skills have a lower mean task performance
after maintenance, with an interval spanning improvement and
decline.

The intervals use 10{,}000 paired bootstrap resamples of
task--change pairs on TB-2, tasks on AppWorld, and variations
on ScienceWorld. Each resample keeps the policies' outcomes
for the same evaluation task together. IRS results average
five forms with ten evaluation applicants each. Two
independent Haiku repair runs differ by .015 in their
five-form mean and by up to .08 on an individual form.

\par\medskip\noindent
\begin{minipage}{\columnwidth}
\centering
\captionof{table}{\textbf{Selective minus No maintenance.}
Both columns are in percentage points, with 95\% paired-bootstrap
intervals. Cost reduction uses the no-skill $W_1$ cost as its
denominator. The paired units are 38 TB-2 task--change pairs,
51 AppWorld tasks, and 41 ScienceWorld variations.}
\label{tab:maint-paired}
\label{tab:maintenance-statistics}
\scriptsize
\setlength{\tabcolsep}{4pt}
\renewcommand{\arraystretch}{1.05}
\begin{tabular}{@{}lcc@{}}
\toprule
& \textbf{$\Delta$Performance} & \shortstack{\textbf{$\Delta$Cost}\\\textbf{reduction}} \\
\midrule
\multicolumn{3}{@{}l}{\tb} \\
\quad Opus, cache   & $+10.5$ $[+2.2,+20.6]$ & $+12.4$ $[+6.6,+20.9]$ \\
\quad Opus, proc.   & $-0.9$ $[-2.6,0.0]$    & $+26.4$ $[+2.1,+56.0]$ \\
\quad Haiku, cache  & $+8.8$ $[+2.6,+17.1]$  & $+10.0$ $[+3.4,+19.6]$ \\
\quad Haiku, proc.  & $+8.8$ $[+0.9,+18.4]$  & $+1.2$ $[-6.5,+11.8]$ \\
\addlinespace[2pt]
\multicolumn{3}{@{}l}{\appworld} \\
\quad Haiku, cache  & $+3.9$ $[-9.8,+17.6]$  & $+16.4$ $[+6.5,+26.2]$ \\
\quad Haiku, proc.  & $-5.9$ $[-21.6,+9.8]$  & $+2.8$ $[-6.4,+11.7]$ \\
\addlinespace[2pt]
\multicolumn{3}{@{}l}{\scienceworld} \\
\quad Haiku, cache  & $0.0$                  & $+17.1$ $[+14.1,+20.4]$ \\
\quad Haiku, proc.  & $0.0$                  & $+17.3$ $[+12.3,+22.8]$ \\
\bottomrule
\end{tabular}
\end{minipage}
\par\medskip

\paragraph{When rebuilding helps.}
Repair and rebuilding offer different performance--cost
tradeoffs. For Opus with procedural IRS skills, rebuilding
uses fewer turns while repair achieves higher accuracy.
Selective follows the validation decisions, so its aggregate
result can differ from both standalone policies.

Most maintenance ends with repair. Across the four TB-2
model--style combinations, maintenance is triggered 55 times:
40 repairs and nine rebuilds are accepted, while six skills
are retained. On IRS, 17 of the 20 form--model--style
combinations accept a repair, twelve on the first attempt;
the other three accept a rebuilt procedural skill.

The procedural ScienceWorld example shows the role of
validation. Its repair is rejected, and rebuilding produces
an accepted update. On the evaluation variations, the rejected
repair scores .561, while the rebuilt skill scores 1.000
with 16.1 steps. The acceptance decision uses validation
results; these evaluation outcomes describe its consequence.

\subsection{Cost and Amortization}
\label{app:cost-accounting}
\label{app:repair-vs-reinduce}
\label{app:monitoring-cost}
\label{app:token-amortization}
\label{app:static-amortization}

\paragraph{Replay.}
Replay incurs environment work whenever the checks run,
including when no update is needed. It makes no model calls.
Table~\ref{tab:replay-timing} reports complete replay runtimes,
measured in a single process on the environment versions
used in the experiments. We time each IRS skill once
(five forms, two styles), each AppWorld task under both
styles and versions, and each ScienceWorld variation under
both styles. Container setup and execution for TB-2 and
validation environment compute are outside these timings.

\par\medskip\noindent
\begin{minipage}{\columnwidth}
\centering
\captionof{table}{\textbf{Elapsed time for a complete replay.}
Times are seconds. Median entries with two endpoints summarize
the range of medians across the recorded timing groups; the
observed range spans individual invocations.}
\label{tab:replay-timing}
\scriptsize
\setlength{\tabcolsep}{5pt}
\renewcommand{\arraystretch}{1.08}
\begin{tabular}{@{}lrr@{}}
\toprule
\textbf{Environment} & \textbf{Median} & \textbf{Observed range} \\
\midrule
\irs & 1.2 & 0.9--1.8 \\
\appworld & 0.2--0.3 & 0.1--3.5 \\
\scienceworld & 3.6--4.3 & 2.6--11.8 \\
\bottomrule
\end{tabular}
\end{minipage}
\par\medskip

\paragraph{Updating a skill.}
On IRS, repair consumes 58.6--142.8K model tokens per
maintenance event, compared with 133.6--197.9K for rebuilding.
These totals include verification, candidate generation,
and acceptance validation. Verification contributes about
19--20K tokens to either operation. Rebuilding consumes
1.24--2.28 times the corresponding repair total, supporting
the choice to try repair first.

\paragraph{Future reuse.}
The token savings from subsequent tasks can repay this
initial cost. Table~\ref{tab:maintenance-amortization} compares
maintenance with retiring the skill, giving an estimated
payback of 21--42 uses. For any baseline $b$, the calculation is
\[
 n_b^\star=\frac{K_{\mathrm{sel}}}{G_b-G_{\mathrm{sel}}},
 \qquad G_b>G_{\mathrm{sel}},
\]
where $K_{\mathrm{sel}}$ counts maintenance steps along
Selective's decision path, and $G_b$ and $G_{\mathrm{sel}}$
are tokens per subsequent task. Maintenance expenditure
excludes reported evaluation runs.

Continuing to use the stale skill gives a different cost
comparison, because it may already be inexpensive even when
its performance is poor. Against this baseline, estimated
payback takes 39--105 uses. Across the 20 IRS
form--model--style combinations, maintenance saves tokens
per use in 18 compared with retirement and in 15 compared
with continued stale reuse. A nonpositive saving gives no
finite token payback, although maintenance may still improve
task performance.

\end{document}